\PassOptionsToPackage{nomath}{stix}
\PassOptionsToPackage{hyphens}{url}
\documentclass[a4paper,fleqn]{cas-dc}

\usepackage[authoryear]{natbib}
\usepackage{microtype}
\makeatletter
\let\casorig@Hy@Warning\Hy@Warning
\def\Hy@Warning#1{\in@{empty anchor}{#1}\ifin@\else\casorig@Hy@Warning{#1}\fi}
\makeatother

\usepackage{amssymb}
\usepackage{amsmath}
\usepackage{array}
\usepackage{booktabs}
\usepackage{tabularx}
\usepackage{threeparttable}
\usepackage{tikz}
\usepackage[export]{adjustbox}
\usepackage{float}
\usepackage{caption}
\usetikzlibrary{arrows.meta,positioning}

\makeatletter
\renewcommand\subsubsection{\@startsection{subsubsection}{3}{\z@}%
  {3.25ex\@plus 1ex \@minus .2ex}%
  {1.5ex \@plus .2ex}%
  {\normalfont\normalsize\bfseries}}
\makeatother

\begin{document}
\let\WriteBookmarks\relax
\renewcommand{\floatpagefraction}{0.5}
\renewcommand{\textfraction}{0.15}
\renewcommand{\topfraction}{0.85}
\renewcommand{\bottomfraction}{0.7}

\shorttitle{BathyFacto for Photogrammetric Bathymetry}
\shortauthors{Brezovsky et al.}

\title[mode=title]{BathyFacto: Refraction-Aware Two-Media Neural Radiance Fields for Bathymetry}

\author[inst1]{Markus Brezovsky}[orcid=0009-0002-3620-1974]
\cormark[1]
\ead{markus.brezovsky@tuwien.ac.at}
\credit{Conceptualization, Methodology, Software, Validation, Writing - Original Draft}

\author[inst2]{Anatol G\"unthner}
\credit{Methodology, Visualization, Writing - Review \& Editing}

\author[inst3]{Frederik Schulte}
\credit{Methodology, Validation, Writing - Review \& Editing}

\author[inst3]{Lukas Winiwarter}
\credit{Methodology, Validation, Funding acquisition, Writing - Review \& Editing}

\author[inst2]{Boris Jutzi}
\credit{Supervision, Resources, Funding acquisition, Writing - Review \& Editing}

\author[inst1]{Gottfried Mandlburger}
\credit{Conceptualization, Supervision, Funding acquisition, Writing - Review \& Editing}

\affiliation[inst1]{organization={Department of Geodesy and Geoinformation, TU Wien},
            city={Vienna},
            country={Austria}}
\affiliation[inst2]{organization={Institute of Photogrammetry and Remote Sensing (IPF), Karlsruhe Institute of Technology (KIT)},
            city={Karlsruhe},
            country={Germany}}
\affiliation[inst3]{organization={Unit of Geometry and Surveying, University of Innsbruck},
            city={Innsbruck},
            country={Austria}}

\cortext[1]{Corresponding author}

\begin{abstract}
Through-water photogrammetry based on UAV imagery enables shallow-water bathymetry, but refraction at the air-water interface violates the straight-ray assumption of Structure-from-Motion and causes systematic depth bias. We present BathyFacto, a refraction-aware two-media extension of Nerfacto integrated into Nerfstudio that targets metrically consistent underwater point clouds on simulated data. BathyFacto uses a shared hash-grid-based density field with a medium-conditioned color head that receives a one-bit medium flag (air or water) and traces each camera ray as two segments: a straight segment in air up to a planar water surface and a refracted segment in water computed via Snell's law with known refractive indices. To allocate samples efficiently across the air--water boundary, we employ a single proposal-network sampler that operates on a virtual straight ray spanning both media, combined with a kinked density wrapper that transparently corrects water-segment positions along the refracted direction before density evaluation. A data adaptation pipeline converts photogrammetric reconstructions to a Nerfstudio-compatible format, estimates the water plane from boundary markers, and provides per-pixel medium masks to gate refraction. We also extend the point cloud export with refraction-corrected backprojection and reversible coordinate transforms to world and global frames. On a simulated two-media scene with known ground truth, the refraction-aware BathyFacto achieves a Cloud-to-Mesh signed median deviation of $-0.001$\,m and 85.7\,\% completeness at 0.2\,m tolerance, measured in the absolute global coordinate frame without any rigid-body alignment, compared to $+1.370$\,m / 11.6\,\% for the single-medium Nerfacto baseline and $+1.409$\,m / 9.9\,\% for BathyFacto without consideration of refraction. Even after a naive refractive-index depth correction, these baselines remain offset by $\approx 0.4$\,m. In contrast to a refraction-corrected Multi-View Stereo reference, which is reliable only for near-nadir viewing geometry, BathyFacto recovers consistent geometry across the full range of camera incidence angles.
\end{abstract}

\begin{highlights}
\item Refraction-aware two-media NeRF for photogrammetric bathymetry.
\item Explicit two-segment ray tracing with Snell's-law refraction at the water surface.
\item Single-sampler architecture with kinked density for air--water sample allocation.
\item Refraction-corrected point cloud export in Nerfstudio for metric bathymetry.
\end{highlights}


\maketitle

\section{Introduction}
\label{sec:introduction}

Shallow-water bathymetry is essential for, among others, hydrodynamic-numerical modeling, flood risk assessment, habitat mapping, and infrastructure planning along rivers, lakes, and coastal zones. Unoccupied aerial vehicles (UAVs) equipped with consumer-grade cameras offer a cost-effective and flexible platform to acquire high-resolution imagery of such environments \citep{westoby2012sfm}. Structure-from-Motion (SfM) and Multi-View Stereo (MVS) photogrammetry can then recover dense 3D point clouds from these images, enabling rapid survey workflows \citep{schoenberger_2016sfm,hirschmueller_2008}. This has also been explored for shallow-water mapping when imaging conditions are suitable \citep{dietrich2017bathymetric,delsavio2023_Using}.

However, through-water photogrammetry faces a fundamental challenge: light rays refract at the air-water interface according to Snell's law, bending toward the surface normal as they enter the denser medium. Standard SfM-MVS pipelines assume straight ray paths and therefore produce systematically biased underwater geometry reconstructions in multimedia settings \citep{maas2015two}. Although post-processing corrections based on known water surface geometry and viewing angles can reduce this bias \citep{agrafiotis2019correcting, kastner2023iterative}, they require handling of each image ray and remain sensitive to uncertainties in surface estimation.

Active sensing with airborne laser bathymetry (ALB) provides an alternative reference by measuring range through water using a pulsed green laser that delivers depth measurements in suitable conditions. However, it is expensive, requires specialized sensors and is limited by legal, environmental and surface conditions \citep{mandlburger2022_review}. Passive, image-based methods therefore remain attractive for routine monitoring, provided refraction can be modeled adequately.

Neural Radiance Fields (NeRFs) enable novel-view synthesis by representing scenes as continuous volumetric functions learned from posed images \citep{mildenhall2020nerf}. Extensions such as Mip-NeRF~360 \citep{barron2022mipnerf360} handle unbounded scenes and anti-aliasing, while Nerfstudio \citep{tancik2023nerfstudio} provides a modular framework that supports research and deployment. Complementary advances in 3D Gaussian Splatting \citep{kerbl2023gaussiansplatting} achieve real-time rendering, and underwater-focused methods like SeaThru-NeRF \citep{levy2023seathrunerf} address scattering and color restoration. However, all of these NeRF variants assume a single homogeneous medium and do not model refraction at interfaces.

NeRFrac \citep{zhan2023nerfrac} explicitly incorporates a refractive surface into the NeRF formulation. It learns both the scene radiance field and the distance to a potentially wavy, refractive interface, enabling novel-view synthesis of underwater scenes observed through water. While NeRFrac reports qualitative results, it focuses on rendering quality rather than geometric accuracy and does not provide a pipeline for exporting metrically meaningful point clouds suitable for bathymetric applications. Moreover, its Vanilla-NeRF backbone does not incorporate modern sampling strategies such as proposal networks, which can limit convergence speed and stability on datasets with large viewpoint variation.

\vspace{4mm}
In our contribution, we present \textbf{BathyFacto}, a refraction-aware two-media extension of Nerfacto specifically designed for photogrammetric bathymetry. BathyFacto is integrated into Nerfstudio and targets improved geometric accuracy of underwater point clouds with refraction-corrected export in reproducible coordinate frames. Our main contributions are as follows:

\textbf{Two-media Nerfacto with ray refraction at the air-water interface based on Snell's law.} We extend Nerfacto with a shared hash-grid-based density field and a medium-conditioned color head that receives a one-bit medium flag (air or water). Each camera ray is traced as two piecewise linear segments, first in air and a second in water, refracted at the air-water interface via Snell's law with known refractive indices.

\textbf{Single-sampler architecture with kinked density.} Rather than maintaining separate sampling hierarchies for air and water, we employ a single proposal-network sampler that operates on a virtual straight ray spanning both media. A kinked density wrapper transparently corrects water-segment positions along the refracted direction before density evaluation, enabling the proposal network to adaptively allocate samples across the air--water boundary.

\textbf{Nerfstudio integration.} We implement BathyFacto within the Nerfstudio framework, leveraging its hash-grid encoding, proposal-network sampling, camera pose optimization, and training infrastructure. The accompanying data pipeline supports per-pixel medium masks for mask-gated two-media rendering and export. This integration promotes reproducibility and allows practitioners to benefit from ongoing Nerfstudio improvements.

\textbf{Refraction-corrected point cloud export.} We extend the standard NeRF point cloud exporter to back-project depth predictions along the refracted two-segment ray geometry. Reversible coordinate transformations allow for exporting in a normalized global reference frame for direct comparison with reference geometry (e.g., simulation ground truth).

\textbf{Ablation study.} We evaluate BathyFacto on a simulated two-media scene with known ground truth, reporting 3D point cloud metrics. We compare configurations with and without refraction correction to quantify the benefit of explicit two-media modeling.

\vspace{4mm}
The remainder of this contribution is organized as follows.
Section~\ref{sec:related-work} reviews related work on through-water photogrammetry, standard NeRFs, and refractive NeRFs. Section~\ref{sec:methods} describes the BathyFacto methodology, starting with the adaptation of the data and continuing to the shared-geometry two-media architecture, interface coupling, and refraction-corrected point cloud export. Section~\ref{sec:experimental-setup} details the simulation dataset providing ground truth, describes the hyperparameters and model variants (Nerfacto baseline, BathyFacto with and without refraction), and outlines the evaluation metrics.
The results derived by BathyFacto are presented in Section~\ref{sec:results}.
In Section~\ref{sec:discussion} our proposed method is discussed and the practical strengths and limitations are summarized. With Section~\ref{sec:conclusion} the contribution concludes and an outlook to future research is provided.

\section{Related Work}
\label{sec:related-work}

This section situates BathyFacto within prior work on optical bathymetry and through-water photogrammetry, standard NeRF reconstruction, and NeRF variants that explicitly model refraction. We first summarize established correction strategies and sensing modalities in multimedia settings, then review NeRF foundations and recent refractive extensions to motivate the need for explicit two-media ray modeling in reconstructions.

\subsection{Through-Water Photogrammetry}
UAV-based photogrammetry provides a flexible and cost-effective option for high-resolution topographic mapping \citep{westoby2012sfm}. Whenever image rays traverse more than one medium with different refractive indices, the interface violates the straight-ray assumption underlying standard SfM and MVS, because rays refract according to Snell's law as they enter the denser medium \citep{delsavio2023_Using}. In photogrammetric terms, the collinearity condition is not met in multimedia settings, and ignoring refraction introduces systematic geometric errors \citep{maas2015two}.

Passive correction strategies typically estimate the water surface geometry and then correct underwater rays either as a post-processing step or within the optimization. Image-/ray-space correction approaches reproject underwater observations using a planar or locally estimated interface and camera geometry \citep{agrafiotis2019correcting}. Iterative refraction-correction workflows apply Snell's law together with camera viewing geometry to adjust SfM-MVS-derived bathymetric point clouds and reduce depth bias \citep{kastner2023iterative}. In practice, the achievable accuracy depends on the validity of the assumed interface model and the quality of the image texture in water \citep{maas2015two, agrafiotis2019correcting}, which is often hampered by sediment-induced turbidity.

Active sensing provides an alternative reference standard. Airborne laser bathymetry (ALB) directly measures range through water using a pulsed green laser and post-processing properly accounts for two-media propagation effects \citep{guenther_2000,mandlburger2022_review}. Although ALB provides depth measurements under suitable conditions, it requires specialized sensors and flight operations. For UAV-based image surveys, a remaining challenge is to couple a physically correct two-media image formation model with 3D reconstruction pipelines such that the exported underwater geometry is metrically interpretable.

\subsection{Photogrammetric Workflow}
The estimation of intrinsic and extrinsic camera orientations is a fundamental prerequisite for downstream 3D reconstruction frameworks, such as MVS and NeRFs. This requirement is traditionally addressed through SfM, a photogrammetric workflow that reconstructs 3D scenes from 2D imagery by simultaneously estimating camera poses and a sparse tie point cloud \citep{schoenberger_2016sfm}. The process is initiated by extracting unique image features, typically via the Scale Invariant Feature Transform (SIFT) algorithm \citep{lowe_2004}. Subsequently, these descriptors are matched in stereoscopic image pairs using robust estimation techniques, such as Random Sample Consensus (RANSAC) \citep{fischler_1981}, to filter out outliers. Following the derivation of relative camera poses from these matches, the pipeline culminates in a global bundle adjustment \citep{triggs_2000}. This optimization refines the interior and exterior camera parameters and 3D point locations simultaneously by minimizing the total reprojection error, thereby ensuring the geometric fidelity of the calibration and the resulting point cloud. MVS constitutes a framework that builds upon the SfM workflow to derive a dense point cloud by leveraging sparse tie points in conjunction with the intrinsic and extrinsic camera orientation parameters \citep{schoenberger_2016mvs,hirschmueller_2008}. The MVS process starts with the detection of corresponding image points constrained by epipolar geometry, subsequently transforming these correspondences into disparity maps which are then converted into per-pixel depth information using the known camera parameters. To generate a comprehensive and dense 3D representation of the scene, these individual depth maps are fused and integrated \citep{furukawa_2015}. On well-textured surfaces, these algorithms exhibit highly accurate surface reconstructions. Conversely, the fidelity of the reconstruction decreases drastically or may fail entirely under suboptimal illumination, particularly on textureless or refractive surfaces, such as those encountered in bathymetric scenarios.

\subsection{Neural Radiance Fields}
NeRFs model scenes as continuous volumetric functions that map 3D position and viewing direction to density and radiance, enabling high-quality novel-view synthesis from oriented images \citep{mildenhall2020nerf}. For large-scale captures, extensions such as Mip-NeRF~360 \citep{barron2022mipnerf360} improve sampling and stability in unbounded scenes, and widely used frameworks such as Nerfstudio with its Nerfacto variant standardize training, evaluation, and deployment \citep{tancik2023nerfstudio}. In parallel, 3D Gaussian Splatting provides a complementary point-based radiance field representation with real-time rendering performance \citep{kerbl2023gaussiansplatting}. These developments have improved reconstruction quality and usability; however, most NeRF variants still assume straight rays in a single homogeneous medium.

For photogrammetric bathymetry, this assumption is problematic: the refractive interface is often a dominant error source, not only the representational capacity of the radiance field. Consequently, the key requirement is not merely plausible view synthesis but a reconstruction pipeline that maintains geometric fidelity under two-media ray bending and supports metric 3D geometry export, both for comparison to external references and for applications that explicitly target underwater geometry rather than only image quality. In the following, we therefore review refraction-aware NeRF approaches.

\subsection{Refractive NeRFs}
Recent NeRF literature addresses non-Lambertian effects and refractive phenomena from different perspectives. Ref-NeRF improves view synthesis of glossy surfaces by explicitly parameterizing view-dependent radiance via reflected directions \citep{verbin2022refnerf}, and REF$^2$-NeRF extends this direction to jointly model reflection and refraction in NeRF-style rendering \citep{kim2023ref2nerf}. While these models improve appearance under complex light transport, they primarily target photorealistic rendering and often rely on implicit or simplified geometric mechanisms rather than physically-grounded explicit two-media ray tracing.

More directly related to BathyFacto are approaches that incorporate ray bending. LB-NeRF addresses light bending in transparent media within a NeRF-style formulation by learning an offset field that displaces sampling points away from the straight camera ray \citep{fujitomi2022lbnerf}. This implicitly represents refraction without directly enforcing Snell's law or an explicit interface geometry: a separate network predicts per-sample 3D offsets as a function of position and view direction, regularized to limit ambiguity in the learned offsets. While this makes the method lightweight and independent of prior knowledge on refractive indices or surface shape, it also decouples ray bending from a physically constrained two-media model and is demonstrated on object-centric scenes with compact transparent volumes rather than extended air-water interfaces typical of UAV bathymetry.
NeRFrac explicitly embeds Snell's law into the NeRF pipeline by modeling a refractive interface of planar or smoothly varying shape and tracing rays as two segments, with joint optimization of interface shape and radiance field \citep{zhan2023nerfrac}. The method is evaluated on synthetic underwater scenes generated by ray tracing and on real data captured with a fixed multi-camera array above a water tank, showing that physically based refraction modeling improves novel-view synthesis and enables reconstruction of the water surface itself. However, the focus remains on image-space quality. The approach is not designed for point cloud export and has not been evaluated on large-scale UAV imagery.

A subsequent extension refines this method by improving refractive surface reconstruction and rendering quality through ray-wise Bezier surface fitting of the interface, adaptive hierarchical sampling of refracted rays, and a hybrid directional encoding based on spherical harmonics \citep{zhang2026refractive}. The evaluation nevertheless remains restricted to small-scale synthetic and laboratory water tank setups without point cloud export, and thus does not address 3D scene reconstruction from real-world UAV imagery.

For UAV-based bathymetry, NeRFrac has first been applied to outdoor river scenes by training on nadir UAV imagery over a pre-Alpine River and evaluating its refractive modeling exclusively with 2D image-space metrics, without exporting or validating 3D point clouds \citep{gunthner2025_NeRFracApplicationToUAV}. Subsequent work extends NeRFrac with a dedicated point cloud export and qualitatively inspects refractive 3D reconstruction from UAV data, but the resulting geometry remains in a normalized internal coordinate system and has not yet been metrically validated against external references \citep{brezovsky2025_Analysis}.

In summary, existing refractive NeRFs demonstrate the benefits of explicit refraction modeling on synthetic and laboratory data, but they either rely on implicit offset fields without physical constraints or are tailored to controlled camera arrays and do not yet provide a workflow for metrically interpretable 3D reconstruction from UAV imagery with metric point cloud export. BathyFacto addresses this gap.

\section{Methodology}
\label{sec:methods}
This section describes the BathyFacto pipeline which is depicted in \autoref{fig:bathyfacto-workflow}. Our workflow consists of three stages: i)~converting photogrammetric outputs into a Nerfstudio-compatible dataset, including water-surface metadata, as detailed in Section~\ref{sec:methods:dataset}; ii) training BathyFacto as a two-media Nerfacto extension that renders each camera ray with a straight segment in air and, where applicable, a refracted segment in water, described in Sections~\ref{sec:methods:architecture}-\ref{sec:methods:losses}; and iii) exporting a refraction-corrected point cloud via two-segment backprojection in reversible coordinate frames, outlined in Section~\ref{sec:methods:export}.

\vspace{0.5em}
\begin{minipage}{\columnwidth}
    \centering
    \begin{tikzpicture}[flowbox/.style={draw, rounded corners, align=left, inner sep=3pt, text width=0.95\columnwidth, font=\scriptsize},
      arrow/.style={-Latex, thick},
      node distance=3mm and 0mm]
      \node[flowbox] (sfm) {%
        \textbf{Photogrammetry (SfM)} \hfill (e.g., Agisoft Metashape)\\
        outputs: oriented images, camera calibration/poses, interface markers};
      \node[flowbox, below=of sfm] (prep) {%
        \textbf{Data Adaptation} \hfill (automated)\\
        fit planar water surface; normalize coordinates; define scene box\\
        write \texttt{train.npz}, \texttt{val.npz}, \texttt{data.npz}};
      \node[flowbox, below=of prep] (dp) {%
        \textbf{BathyNerfDataParser} \hfill (Nerfstudio)\\
        reads \texttt{train.npz}/\texttt{val.npz} for training and \texttt{data.npz} for export\\
        loads cameras, scene-box (AABB) bounds, water-plane metadata, and masks};
      \node[flowbox, below=of dp] (train) {%
        \textbf{Training} \hfill (Nerfstudio)\\
        two-media rendering with Snell refraction (mask-gated)\\
        output: model checkpoint and configuration snapshot};
      \node[flowbox, below=of train] (export) {%
        \textbf{Point cloud export} \hfill (Nerfstudio)\\
        refraction-corrected backprojection along two-segment rays\\
        output: point cloud in the original global frame};
      \draw[arrow] (sfm) -- (prep);
      \draw[arrow] (prep) -- (dp);
      \draw[arrow] (dp) -- (train);
      \draw[arrow] (train) -- (export);
    \end{tikzpicture}
    
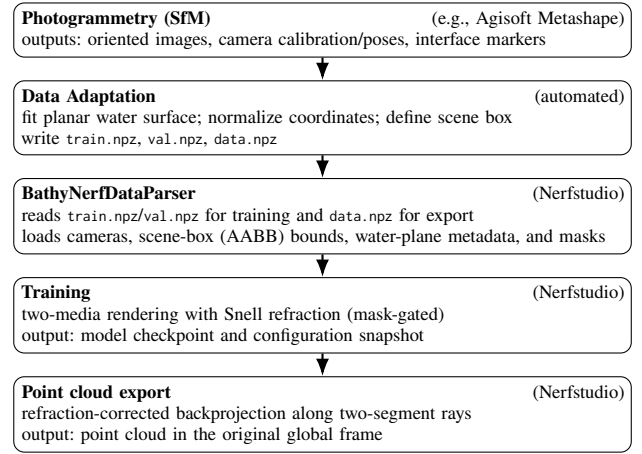
\captionof{figure}{BathyFacto workflow from photogrammetric reconstruction to refraction-corrected point cloud export. Photogrammetry outputs are converted into NumPy archives (\texttt{.npz}). The BathyNerfDataParser loads \texttt{train.npz}/\texttt{val.npz} for training and \texttt{data.npz} for full-scene export. Finally, the refraction-aware exporter back-projects points along the two-segment ray geometry and maps them back to the original global frame.}
    \label{fig:bathyfacto-workflow}
\end{minipage}

\subsection{Data Adaptation}
\label{sec:methods:dataset}
BathyFacto operates on Nerfstudio-style datasets stored as NumPy archives (\texttt{.npz}-files) containing cameras, scene bounds, and auxiliary metadata. Starting from a photogrammetric reconstruction with oriented images, we perform the following steps to obtain these files:

\textbf{Photogrammetric Export.}
We export reconstructed camera intrinsic and extrinsic parameters, as well as 3D markers representing the water surface, which are, for now, placed manually along the land-water boundary. We use Agisoft Metashape and export it as an \texttt{.xml}-file.

\textbf{Pose Transformation.}
The camera poses and 3D marker points are converted from the photogrammetric coordinate system into the OpenCV-style convention used by Nerfstudio.

\textbf{Water Surface Estimation.}
Marker points along the shoreline are used to fit a planar water surface via least-squares adjustment, yielding the plane normal vector and intercept in the photogrammetric frame.

\textbf{Scene Transformation.}
The cameras and markers are first rotated so that the fitted water plane becomes horizontal and the positive $z$-axis points upward towards the cameras. We then translate the scene, shifting the coordinate origin to the centroid of the camera positions, and apply a single global scale factor so that all camera and marker coordinates lie approximately within $[-1,1]$, resulting in numerically stable units in the training space.

\textbf{Scene Box.}
The scene box used for NeRF sampling is derived from the horizontal ($x,y$) extent of the normalized camera positions and the vertical ($z$) range spanned by both cameras and markers. This axis-aligned box is then translated to move its center into the markers' centroid, ensuring that sampling covers both the air and underwater regions in a balanced way. The camera positions, the water surface plane, and the resulting scene box are visualized in \autoref{fig:water-plane-fit}.

\vspace{0.5em}
\begin{minipage}{\linewidth}
    \centering
    \includegraphics[width=1\linewidth]{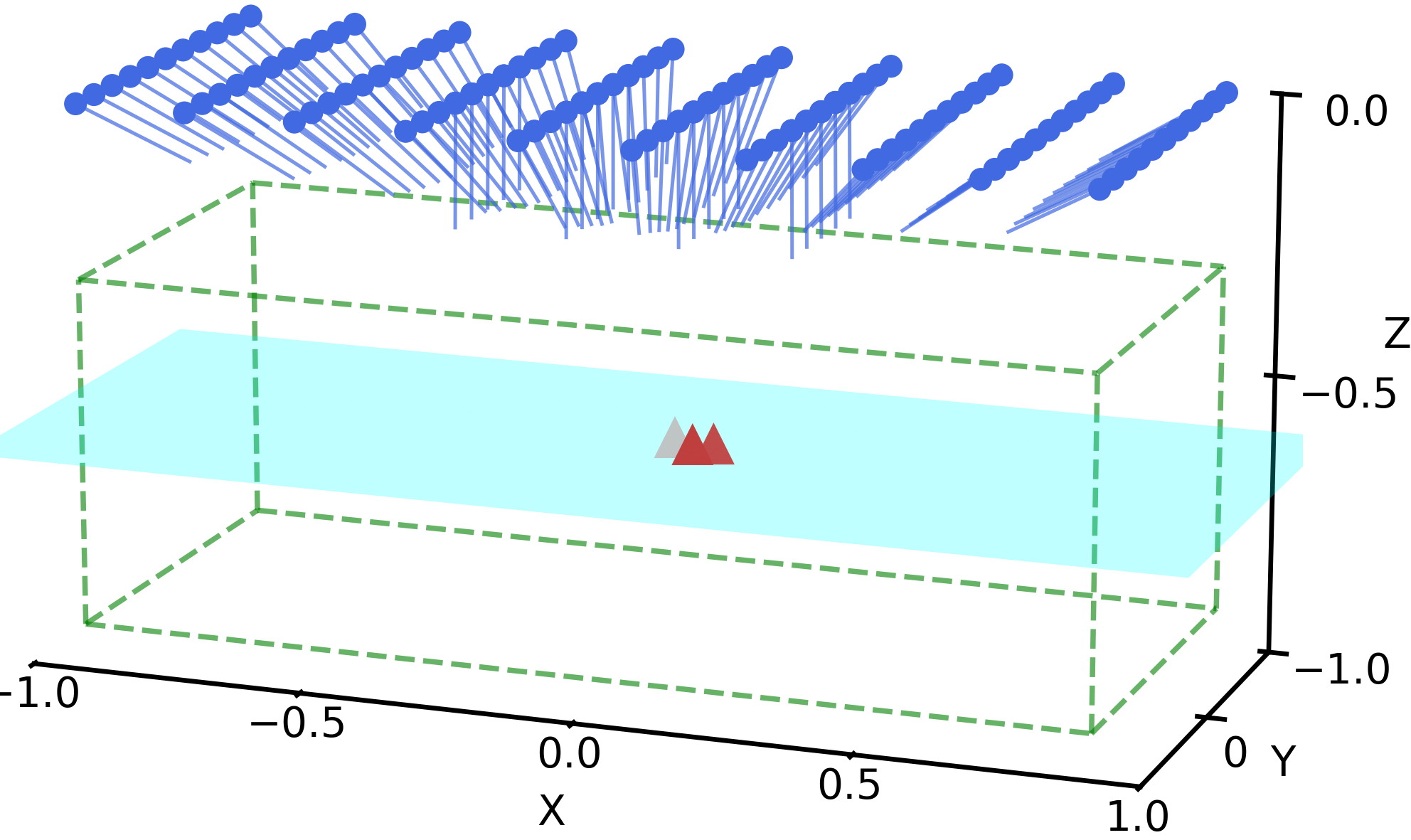}
    \captionof{figure}{Acquisition geometry. Camera positions and viewing directions are shown in blue. Markers (red) define the water surface (cyan), from which the scene box (dashed green) is derived.}
    \label{fig:water-plane-fit}
\end{minipage}
\vspace{0.5em}

\textbf{Medium Masks.}
For mixed land-water scenes, BathyFacto can optionally use per-image masks to restrict refractive rendering to water-covered pixels while keeping land pixels on straight, non-refracted rays. In our synthetic dataset, such masks are available. For other datasets, water masks must be generated in a preprocessing step. Each mask file must share the same filename as its corresponding RGB image, and must be aligned pixel-wise to ensure consistent gating of refractive vs.\ non-refractive rays during training and export. An example water mask for one image is shown in \autoref{fig:medium-mask}.

\textbf{Train/Validation Split and Packaging.}
In a final step, the oriented images are split into a training subset and a validation subset, with nadir and oblique views distributed as evenly as possible across both sets. We use 90\% of all images for training and the remaining images for validation. For each split, the corresponding RGB filenames, medium masks, camera intrinsics, and extrinsics are written to \texttt{train.npz} and \texttt{val.npz}, respectively. In addition, a full \texttt{data.npz} file stores all images, all masks, the complete camera block, and the shared scene metadata, including the water surface parameters, the normalization transform, and the scene box.

\vspace{0.5em}
\begin{minipage}{\linewidth}
\centering
\includegraphics[width=\linewidth]{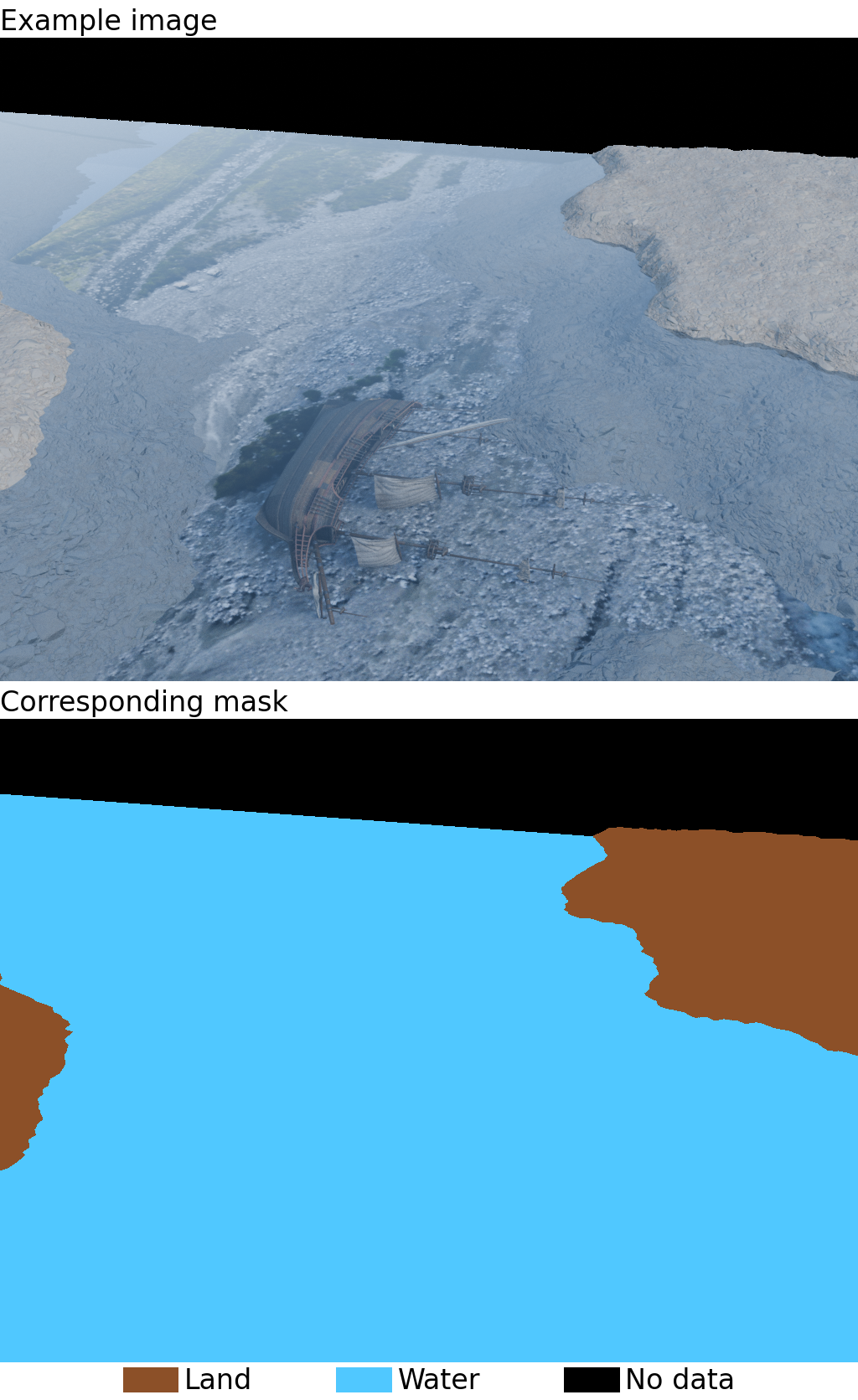}
\captionof{figure}{Example image and corresponding mask used to gate two-media rendering. The mask assigns each pixel to land (air-only), water (two-segment ray with refraction), or ignore/no-sampling regions that are excluded from training losses and evaluation.}
\label{fig:medium-mask}
\end{minipage}

\subsection{BathyFacto Architecture}
\label{sec:methods:architecture}
BathyFacto extends Nerfacto with a shared-geometry two-media field that uses a single density representation for both air and water, combined with a medium-conditioned color head. This design enforces geometric consistency across the air-water interface: the same hash-grid-based density field predicts volume density $\sigma$ everywhere, while a single color MLP receives a one-bit medium flag (0 for air, 1 for water) as additional input to decode medium-specific appearance. Given a sample position $\mathbf{x}$, direction $\mathbf{d}$, and medium flag $m\in\{0,1\}$, the shared field predicts density $\sigma$ and a geometry feature embedding $\mathbf{h}$, which is then passed along with $m$ to the color head to produce a medium-aware RGB color $\mathbf{c}$.

We sample points along rays within the scene box. BathyFacto uses Nerfacto-style proposal-network sampling to focus samples around high-density regions. An Axis Aligned Bounding Box (AABB) near/far collider provides per-ray bounds.

\subsubsection{Shared Field, Encodings, and Physics Parameters}
The main field uses a multi-resolution hash-grid encoding \citep{muller2022instant} with $16$ resolution levels, a base resolution of $16$, a maximum resolution of $2048$, and two features per level, stored in a hash table with a size of $2^{19}$. The density MLP has a hidden dimension of $64$. This encoding replaces the sinusoidal positional encoding used in vanilla NeRF and provides substantially faster convergence and higher spatial fidelity.

The shared density MLP produces both volume density and a geometry feature embedding. A single color head takes the geometry embedding, an encoded viewing direction, optional appearance embeddings, and a one-bit medium flag as input and predicts RGB values. The medium flag (0 for air, 1 for water) is concatenated to the color head input, increasing its input dimension by one. This lightweight conditioning enables the color head to learn medium-specific appearance (such as water absorption and color shift) while sharing all weights across both media, which is more parameter-efficient than maintaining two independent color heads. Optional per-image appearance embeddings (dimension $32$) allow BathyFacto to accommodate photometric variation across the image set.

The two-media physics is controlled by fixed refractive indices $n_{\text{air}}$ and $n_{\text{water}}$ (default: $1.0$ and $1.333$). In the ablation used in this study, we disable refraction at the interface while keeping the same shared-geometry two-media architecture.

\subsubsection{Single-Sampler with Kinked Density}
\label{sec:methods:kinked-density}
Rather than maintaining separate proposal-sampling hierarchies for air and water, BathyFacto uses a single proposal-network sampler that operates on a \emph{virtual straight ray} spanning both media. For each camera ray that intersects the water surface at parameter $t_I$, we construct a virtual ray from the camera near plane to $t_I + t_{\text{far,water}}$, where $t_{\text{far,water}}$ is the maximum water-segment length obtained from the scene box intersection along the refracted direction. The proposal network samples along this virtual straight ray as if it were a single-medium scene.

To ensure that density is evaluated at physically correct positions despite the straight virtual ray, we wrap each proposal density function in a \emph{kinked density} correction. For any sample at parameter $t$ along the virtual ray, the wrapper computes the true 3D position as
\begin{equation}
\mathbf{x}(t) =
\begin{cases}
\mathbf{o} + t\,\mathbf{d} & \text{if } t \le t_I,\\
\mathbf{p}_I + \mathbf{d}_w\,(t - t_I) & \text{if } t > t_I,
\end{cases}
\label{eq:kinked-position}
\end{equation}
where $\mathbf{p}_I = \mathbf{o} + t_I\,\mathbf{d}$ is the interface entry point and $\mathbf{d}_w$ is the refracted direction. The density network receives these corrected positions, so that the proposal distribution reflects the true scene geometry even though the sampler operates on a straight parameterization. This piecewise mapping, defined by Eq.~\eqref{eq:kinked-position}, is also used for the final sample correction after proposal sampling converges: air samples remain on the original ray, while water samples are placed along the refracted direction.

In the experiments conducted in this study, we use two iterations of the proposal network with $(256,96)$ proposal samples and $48$ NeRF samples per ray (Table~\ref{tab:hyperparams}). Volume-rendering weights are computed from the original $t$-values on the virtual straight ray, preserving a consistent transmittance chain from camera to scene bottom.

\subsubsection{Camera Pose Optimization}
BathyFacto supports learnable camera pose refinement via an $\text{SO}(3)\times\mathbb{R}^3$ camera optimizer. During training, the optimizer applies small rotation and translation corrections to each camera's ray bundle, reducing residual pose errors from the upstream SfM calibration. This is especially beneficial for bathymetric scenes where refraction-induced systematic errors may be coupled with pose uncertainty. The optimizer is regularized with L2 penalties on rotation ($\lambda_{\text{rot}}=0.001$) and translation ($\lambda_{\text{trans}}=0.01$).

\subsubsection{Mask-Gated Two-Media Rendering}
Water masks decide whether a pixel is treated as observing water or land. Only rays that both intersect the estimated water surface plane and are flagged by the medium mask extend the virtual ray through the water segment; non-water pixels remain air-only. In the no-refraction ablation, the virtual ray still extends through the interface, but the kinked density wrapper uses the original air direction instead of the refracted direction (refraction disabled). A configurable threshold (default: $0.5$) binarizes the medium mask. Additionally, an optional per-pixel validity mask allows excluding ``no data''-pixels or ignored regions from both training losses and evaluation metrics.

\subsection{Two-media NeRF Interface Coupling}
\label{sec:methods:coupling}
For each camera ray $\mathbf{r}(t)=\mathbf{o}+t\mathbf{d}$, we compute the intersection parameter $t_I$ with the planar water surface. Rays that do not intersect the plane within their near/far range are rendered as single-medium (air) rays. If a water mask indicates that a pixel falls within the water domain and the ray intersects the plane, we compute the interface entry point $\mathbf{p}_I=\mathbf{o}+t_I\mathbf{d}$ and the refracted direction below the surface.

At the interface, we compute the refracted direction $\mathbf{d}_w$ using Snell's law with known refractive indices $n_{\text{air}}$ and $n_{\text{water}}$. Let $\mathbf{n}$ be the unit normal of the planar interface and $\mathbf{d}$ the unit incident direction in air. We first orient $\mathbf{n}$ so that it opposes the incident ray by enforcing
\begin{equation}
\cos\theta_i = -\mathbf{n}^\top\mathbf{d} \ge 0,
\end{equation}
flipping $\mathbf{n}$ when necessary. The direction $\mathbf{d}_w$ of the refracted underwater ray is then calculated by:
\begin{equation}
\mathbf{d}_w=\operatorname{refract}\!\left(\mathbf{d},\mathbf{n},n_{\text{air}},n_{\text{water}}\right).
\end{equation}
The resulting two-segment ray geometry is illustrated in \autoref{fig:two-media-ray}.

\vspace{0.5em}
\begin{minipage}{\linewidth}
    \centering
    \includegraphics[width=1\linewidth]{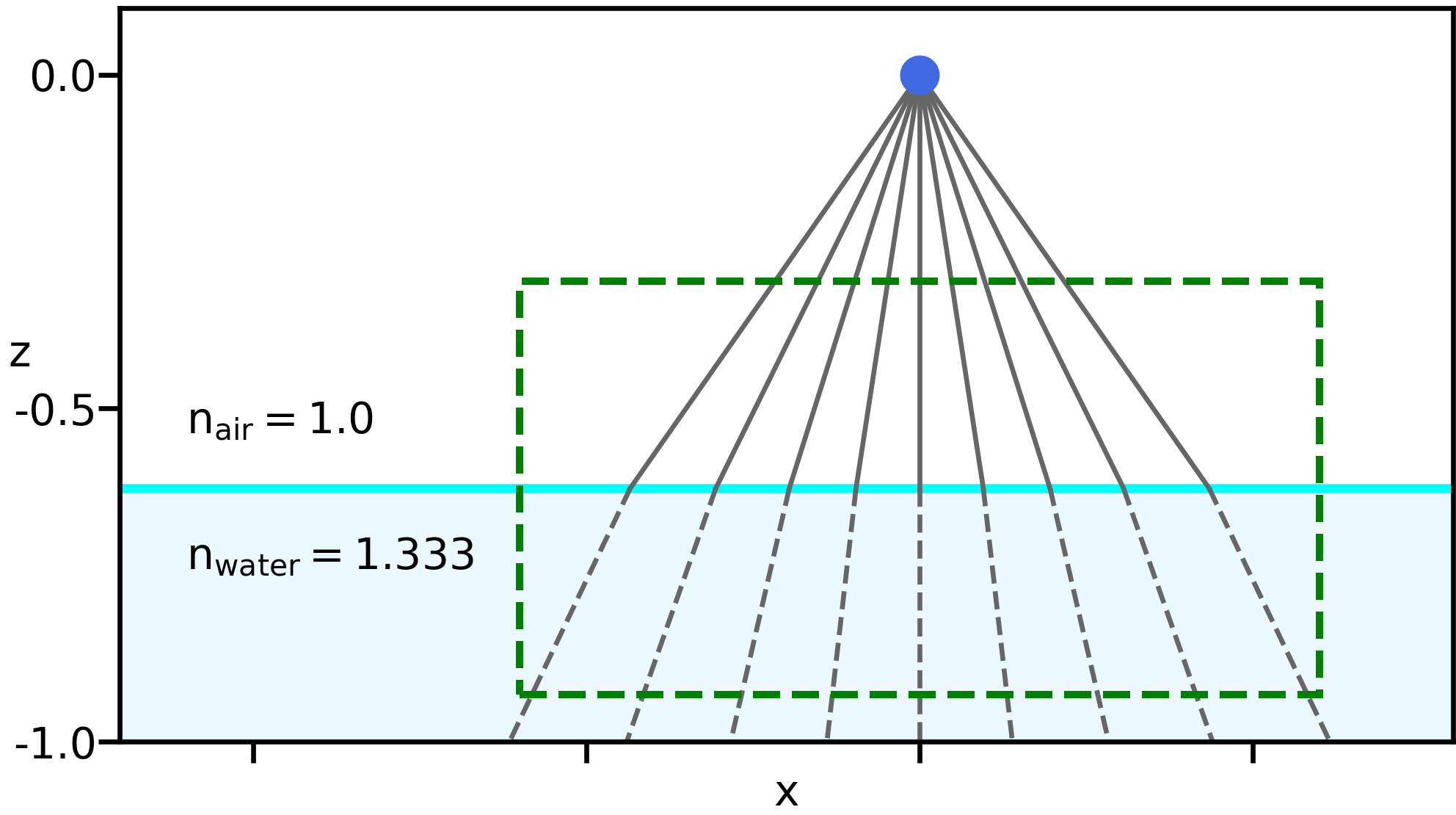}
    \captionof{figure}{Two-segment ray geometry in the BathyFacto coordinate frame. Rays are shown in solid gray, the refracted segments below the water surface as dashed lines. The planar air-water interface separates the two media with their respective refractive indices $\text{n}_{\text{air}} = 1.0$ and $\text{n}_{\text{water}} = 1.333$. The green dashed rectangle denotes the scene box within which the single proposal sampler allocates samples via the kinked density wrapper.}
    \label{fig:two-media-ray}
\end{minipage}
\vspace{0.5em}

We intersect the refracted ray with the scene box to obtain the maximum water-segment length $t_{\text{far,water}}$. A virtual straight ray is then constructed from the camera near plane to $t_I + t_{\text{far,water}}$, providing the single proposal-network sampler with a continuous parameterization across both media.

The kinked density wrapper (Section~\ref{sec:methods:kinked-density}) corrects positions transparently, so the sampler perceives a standard single-medium ray while density is evaluated at physically correct air and water locations.

After proposal sampling, the final NeRF sample positions and viewing directions are corrected analogously. Each sample is assigned a medium flag $m_i$: $m_i=0$ for $t_i \le t_I$ (air) and $m_i=1$ for $t_i > t_I$ (water). The medium flag is passed to the color head for medium-conditioned appearance prediction.

Volume rendering uses the standard alpha-compositing formulation on the $t$-parameterized virtual ray. Given densities $\sigma_i$ and sample intervals $\Delta_i$, we compute
\begin{align}
\alpha_i &= 1 - \exp(-\sigma_i\,\Delta_i),\\
T_1 &= 1,\qquad T_i = T_{i-1}(1-\alpha_{i-1})\ \text{for } i>1,\\
w_i &= T_i\,\alpha_i.
\end{align}
Because the virtual ray provides a monotonically increasing $t$-parameterization across both media, the transmittance chain naturally couples the air and water segments without requiring explicit sample concatenation or sorting. Consequently, underwater samples inherit the residual transmittance of the air segment rather than restarting with unit transmittance at the interface.

\subsection{Deviations from Nerfacto}
\label{sec:methods:differences}
BathyFacto is built on the Nerfacto backbone within Nerfstudio \citep{tancik2023nerfstudio} and retains its hash-grid encoding, proposal-network sampling, and training infrastructure. It differs in three key aspects. First, BathyFacto introduces a physically constrained two-media ray model at a planar air-water interface with explicit Snell's-law refraction (optionally disabled for ablation), whereas Nerfacto assumes a single homogeneous medium. Second, BathyFacto conditions the color head on a one-bit medium flag (air or water) while sharing the density representation, whereas Nerfacto uses a single unified field without medium awareness. Third, BathyFacto wraps the proposal density functions in a kinked density correction that redirects water-segment positions along the refracted ray, enabling a single proposal hierarchy to allocate samples across both media, whereas Nerfacto uses its single proposal hierarchy without position correction. Scene contraction is disabled in all experiments, as the bathymetric scenes are bounded.

\subsection{Training Losses}
\label{sec:methods:losses}
We supervise BathyFacto with a photometric RGB loss between the rendered image and the ground-truth image. For a set of pixels $\mathcal{I}$ with RGB values $\mathbf{C}_i$ and predictions $\hat{\mathbf{C}}_i$, we use an MSE loss,
\begin{equation}
\mathcal{L}_{\text{rgb}}
= \operatorname{mean}_{i\in\mathcal{I}}
\left\lVert \mathbf{C}_i-\hat{\mathbf{C}}_i \right\rVert_2^2.
\end{equation}
If an explicit validity mask is available (e.g., from an alpha channel, ``no data''-pixels, or ignore regions), we exclude invalid pixels and re-normalize by the number of valid pixels to keep the loss magnitude comparable across batches. In the compared BathyFacto runs, medium masks are available for ray gating; the Nerfacto baseline is trained as a single-medium model without this two-media gating. Camera pose optimization adds an additional regularization loss that penalizes excessive rotation and translation corrections.

To reduce ``floater'' artifacts and encourage compact density along rays, we use the distortion regularizer (Mip-NeRF 360 style) on all samples along the virtual ray. When proposal networks are enabled, we additionally apply an interlevel loss that supervises the proposal sampling distribution of the single proposal hierarchy.

Overall, the training objective sums the photometric and regularization terms,
\begin{equation}
\mathcal{L}=\mathcal{L}_{\text{rgb}}+\lambda_{\text{dist}}\mathcal{L}_{\text{dist}}+\lambda_{\text{inter}}\mathcal{L}_{\text{inter}},
\end{equation}
with fixed coefficients ($\lambda_{\text{dist}}=0.002$ and $\lambda_{\text{inter}}=1.0$).

\subsection{Refraction-Corrected Point Cloud Export}
\label{sec:methods:export}
Standard NeRF point cloud export assumes straight rays and therefore produces biased underwater geometry when refraction is present. Therefore, we extend the exporter to back-project points along the two-segment ray geometry: for rays intersecting the air-water interface and reaching into water, we compute the interface point and project the remaining depth along the refracted direction, while non-water rays are projected along the original direction.

In Nerfstudio, the Bathy point cloud exporter samples rays for all dataset images, evaluates the model, and back-projects one 3D point per ray using the predicted export depth. Rays are retained only if their accumulated opacity exceeds a fixed threshold, which suppresses background and other low-confidence samples. When refraction correction is enabled, the exporter uses the learned interface parameters (plane normal and plane intercept) together with the refractive indices to compute the per-ray refraction and applies the two-segment backprojection described above.

Beyond Nerfstudio's standard dataparser transform, the dataset pipeline stores reversible normalization metadata that map points from the normalized training frame back to the original photogrammetric coordinate system. Let $\mathbf{x}_{\text{norm}}$ denote a point in the normalized training frame, $\mathbf{R}_{\text{norm}}$ the rotation that aligns the fitted water-plane normal with the positive $z$-axis, $\mathbf{c}$ the centroid of the rotated camera positions used for centering, and $s_{\text{norm}}$ the isotropic normalization scale. Using column-vector notation, the inverse normalization yields the point in the chunk-local photogrammetric frame:
\begin{equation}
\mathbf{x}_{\text{chunk}}=\mathbf{R}_{\text{norm}}^\top (s_{\text{norm}}\,\mathbf{x}_{\text{norm}}+\mathbf{c}),
\end{equation}
where $\mathbf{x}_{\text{chunk}}$ denotes chunk-local coordinates. Let $s_{\text{chunk}}$, $\mathbf{R}_{\text{chunk}}$, and $\mathbf{t}_{\text{chunk}}$ denote the Metashape chunk transform that maps chunk-local coordinates to the original global frame. The final back-transformation is
\begin{equation}
\mathbf{x}_{\text{global}}=\mathbf{R}_{\text{chunk}}(s_{\text{chunk}}\mathbf{x}_{\text{chunk}})+\mathbf{t}_{\text{chunk}}.
\end{equation}

This enables direct metric comparison with reference geometry in the original global frame and preserves geometric consistency when returning exported point clouds from the normalized training space to the source photogrammetric frame.

\section{Experimental Setup}
\label{sec:experimental-setup}

This section summarizes the data, training setup, and evaluation protocol used in our experiments. We introduce the simulation dataset providing ground truth, describe implementation details and hyperparameters, list the compared configurations, and outline the metrics used to assess image quality and geometric accuracy.

\subsection{Simulation Dataset (Blender)}
The dataset foundational for this study is available for download \citep{schulte_2026data}, with the comprehensive simulation pipeline documented in detail in \cite{schulte_2025simu}. The pipeline starts with the derivation of camera trajectories based on predefined mission parameters, such as image overlap, Ground Sampling Distance, survey area extent, and camera model. In this case, a Point-of-Interest (POI) mission pattern was implemented, generating 130 rendered images along an oblique trajectory. The poses are then imported into Blender (V4.5) \citep{blender_2026} via its Python interface to simulate a photorealistic coastal scene that comprises a ship, terrestrial features, and a subaqueous volume. To simulate underwater conditions, a water block characterized by a refractive index of $n = 1.333$ is implemented, assuming an ideal camera model with zero lens distortion. A significant advantage of this simulation framework is the ability to rapidly iterate physical and mission-contingent variables while maintaining an environment free from uncontrolled systematic biases.

\subsection{Implementation Details \& Hyperparameters}
We evaluated a Nerfacto baseline and BathyFacto in two ablation variants (refraction enabled vs.\ disabled). Table~\ref{tab:hyperparams} summarizes the key parameters. All experiments were run on an NVIDIA L40s GPU (8 CPUs, 32\,GB RAM). BathyFacto (refraction ON) requires approximately 67\,min of training time and 2.4\,GB of GPU memory at 100\,k iterations --- a 40\% increase in training time and 18\% increase in GPU memory over the Nerfacto baseline (48\,min / 2.0\,GB). Point-cloud export (5\,M points) adds approximately 4.5\,min per run.

\begin{table*}[t]
\centering
\begin{threeparttable}
\caption{Key configuration parameters of the compared NeRF models.}
\label{tab:hyperparams}
\small
\setlength{\tabcolsep}{6pt}
\renewcommand{\arraystretch}{1.2}
\begin{tabularx}{\textwidth}{@{}>{\raggedright\arraybackslash}X*{3}{>{\centering\arraybackslash}p{0.20\textwidth}}@{}}
\toprule
\textbf{Parameter} & \textbf{Nerfacto} & \textbf{BathyFacto \newline  (No Refraction)} & \textbf{BathyFacto (Refraction)} \\
\midrule
Max iterations & 100\,000 & 100\,000 & 100\,000 \\
Train rays / batch & 4096 & 4096 & 4096 \\
Eval rays / batch & 4096 & 4096 & 4096 \\
NeRF samples / ray & 48 & 48 & 48 \\
Proposal samples / ray & (256, 96) & (256, 96) & (256, 96) \\
Distortion loss mult. & 0.002 & 0.002 & 0.002 \\
Scene contraction & disabled & disabled & disabled \\
Camera optimizer & None (fixed)\tnote{a} & None (fixed)\tnote{a} & None (fixed)\tnote{a} \\
Proposal schedule (\texttt{warmup/update/anneal}) & 5000 / 5 / on & 5000 / 5 / on & 5000 / 5 / on \\
Proposal anneal max iters & 1000 & 1000 & 1000 \\
Hash-grid levels / max res & 16 / 2048 & 16 / 2048 & 16 / 2048 \\
Hidden dim (density / color) & 64 / 64 & 64 / 64 & 64 / 64 \\
Optimizer & Adam (lr $10^{-2}$) & Adam (lr $10^{-2}$) & Adam (lr $10^{-2}$) \\
Scheduler & Exp.\ decay $\to 10^{-4}$ & Exp.\ decay $\to 10^{-4}$ & Exp.\ decay $\to 10^{-4}$ \\
Refractive indices $(n_{\text{air}}, n_{\text{water}})$ & -- & (1.0, 1.333) & (1.0, 1.333) \\
Media model & single field & shared density, 1-bit conditioned head & shared density, 1-bit conditioned head \\
Refraction model & none (single-medium) & disabled (ablation) & Snell refraction (enabled) \\
\bottomrule
\end{tabularx}
\begin{tablenotes}
\item[a] Camera poses are treated as fixed; no pose optimization is applied, as the simulation dataset provides precise known camera positions.
\end{tablenotes}
\end{threeparttable}
\end{table*}

\subsection{Experimental Configurations}
\label{sec:experimental-setup:configs}
We compare the following configurations to assess the impact of explicit refraction modeling:

\textbf{Baseline (Nerfacto)}: The single-medium Nerfacto configuration used as a standard baseline, with the same hash-grid encoding and training parameters but no two-media modeling.

\textbf{BathyFacto (Refraction)}: Our two-media BathyFacto configuration with planar Snell refraction enabled, shared density field, and medium-conditioned color head.

\textbf{Ablation (BathyFacto No-Refraction)}: The same BathyFacto configuration with refraction disabled (rays continue straight through the interface), used to isolate the benefit of refractive geometry from the shared two-media architecture.

\textbf{MVS baseline}: A refraction-corrected Multi-View Stereo point cloud, included as a reference for the traditional photogrammetric pipeline. Dense points are reconstructed in Agisoft Metashape from spatial image groups (two-image pairs or three-image triplets, each forming its own chunk) and exported, then refraction-corrected in OPALS \citep{pfeifer2014opals} using the known planar water surface and refractive index $n=1.333$. The per-point ray correction follows the through-water dense image matching approach of \citet{mandlburger2019throughwater}. Each chunk is corrected using only its own camera orientations, so that the rays of a group share a consistent incidence geometry.

In all presented experiments, camera poses are treated as fixed (no pose optimization), as the simulation dataset provides precise known camera positions.

\subsection{Evaluation Metrics}
To evaluate the quality of the 3D reconstruction, the assessment procedure described in \cite{schulte_2025simu} was applied. This study has also shown that 2D image-space metrics are not suitable for assessing geometric reconstruction quality.

The generated point clouds and the reference mesh were cropped to the central area of $44 \times 42\,{m}^2$ and filtered with a plane-based statistical outlier removal filter (neighbors = 10, $\sigma= 2.0$) \citep{CloudCompare_2026}. Since point clouds are provided in a common reference frame, distances were evaluated directly in this coordinate system. Using the reference mesh, the cloud-to-mesh (C2M) distances were calculated for all reconstructed point clouds. From these distances, the mean and standard deviation were derived to quantify reconstruction accuracy.

The method from \citet{seitz_2006} was used to estimate completeness, and $15.0$ million points were uniformly sampled across the surface area of the (cropped) reference mesh $\mathcal{P}_{\text{ref}}$, resulting in a denser representation of the target geometry than the reconstructed point clouds. Subsequently, cloud-to-cloud (C2C) distances between $\mathcal{P}_{\text{ref}}$ and each reconstructed point cloud $\mathcal{P}_i$ were computed. Reference points within a distance threshold of $0.2\,\text{m}$ to the nearest reconstructed point were classified as reconstructed ($\mathcal{P}_{\text{rec}}$). Following \citet{hermann2024usegeo}, we adopt $0.2\,\text{m}$ as the established completeness tolerance for UAV-based photogrammetric datasets. Completeness is defined as:
\begin{equation}
\mathrm{Completeness} =
\frac{\lvert \mathcal{P}_{\text{rec}} \rvert}{\lvert \mathcal{P}_{\text{ref}} \rvert}
\end{equation}
\section{Results}
\label{sec:results}

This section presents the qualitative and quantitative results of our evaluation on the simulation dataset.

\begin{figure*}[p]
\centering
\setlength{\tabcolsep}{2pt}
\renewcommand{\arraystretch}{1.0}
\begin{tabular}{@{}r@{\hspace{4pt}}cccc@{}}
& \small\textbf{Ground Truth} & \small\textbf{BathyFacto (Refr.\ OFF)} & \small\textbf{BathyFacto (Refr.\ ON)} & \small\textbf{Nerfacto} \\[3pt]
\rotatebox{90}{\small\textbf{RGB\,(0021)}} &
\includegraphics[width=0.212\textwidth]{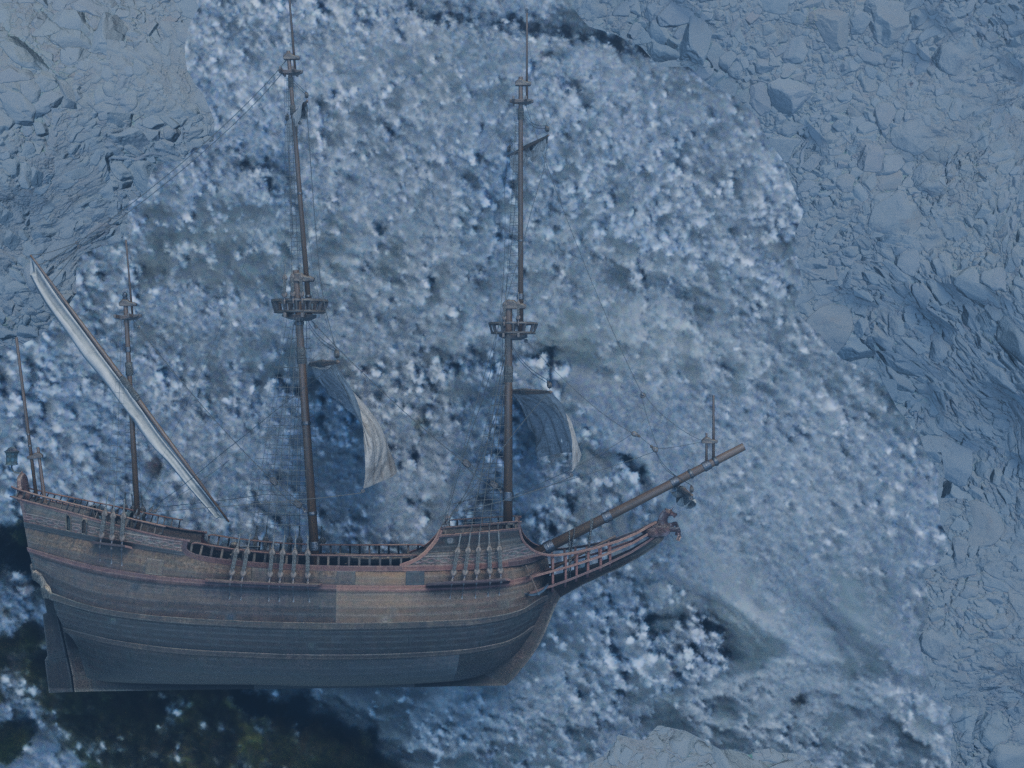} &
\includegraphics[width=0.212\textwidth]{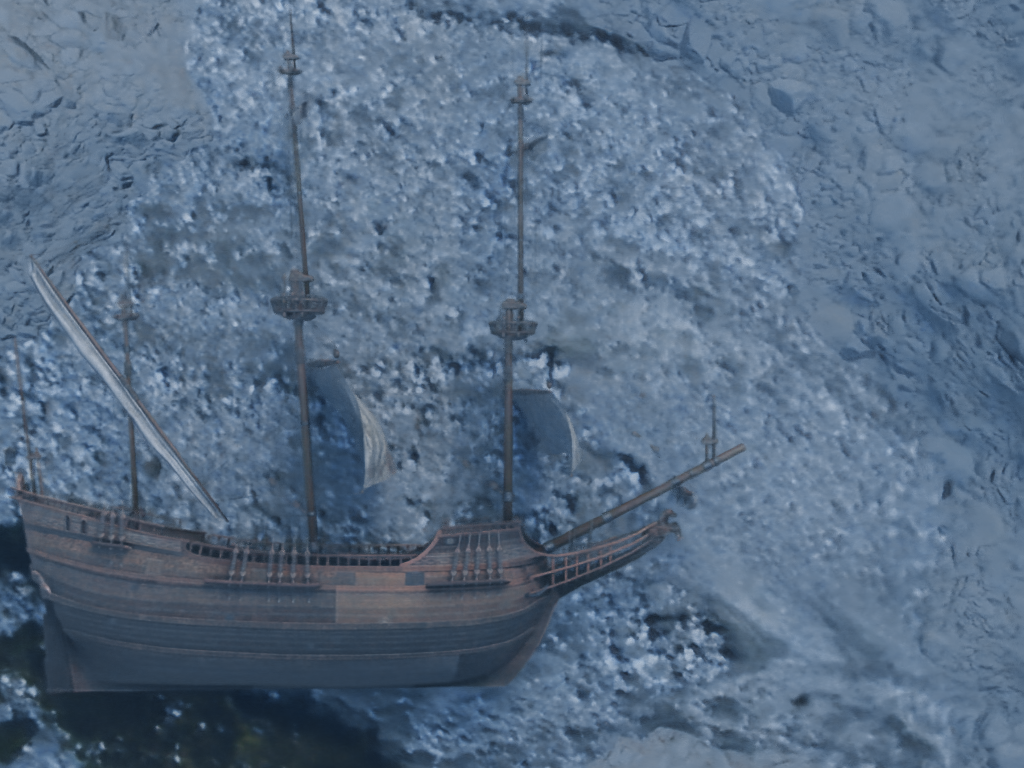} &
\includegraphics[width=0.212\textwidth]{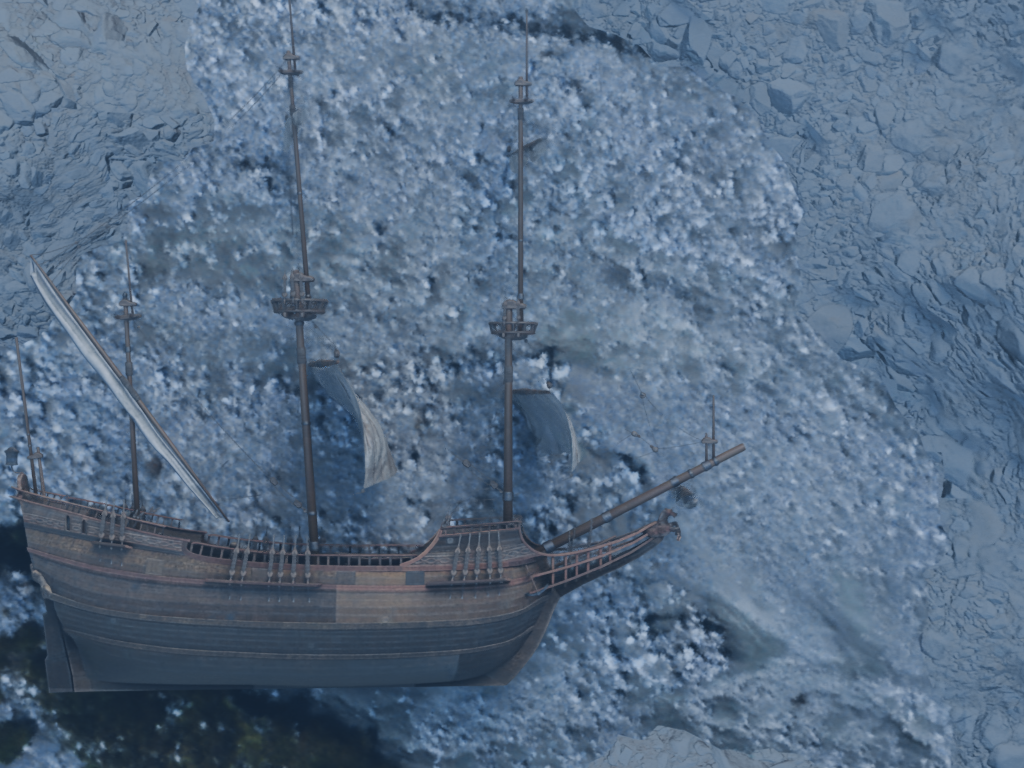} &
\includegraphics[width=0.212\textwidth]{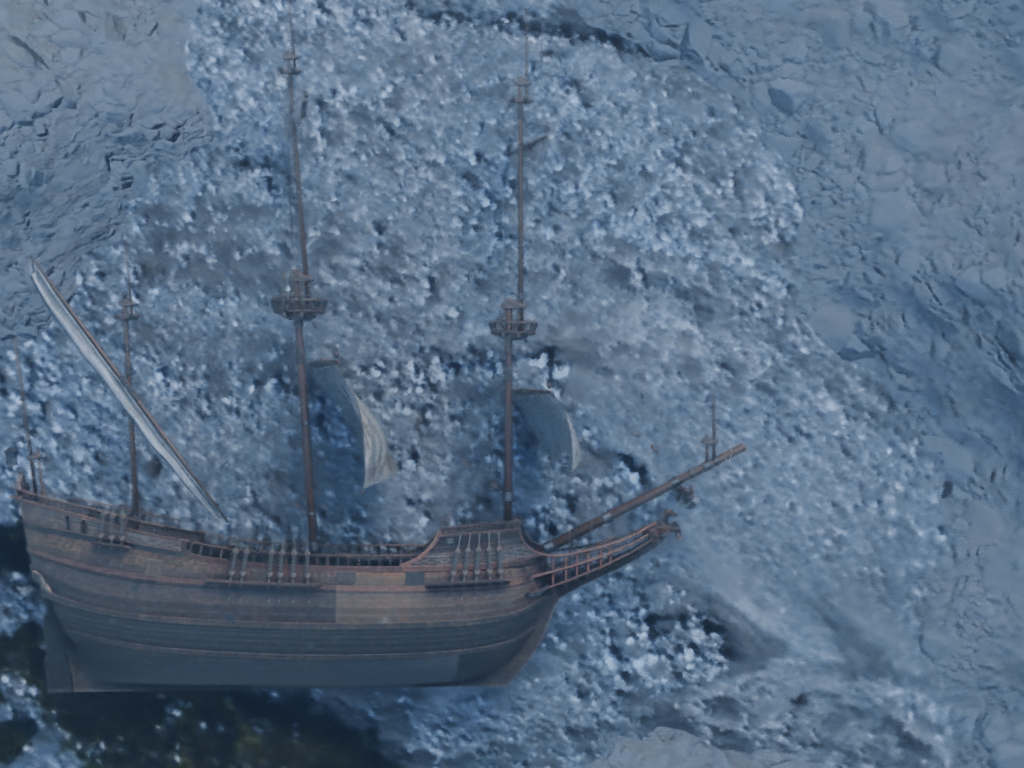} \\[2pt]
\rotatebox{90}{\small\textbf{Depth\,(0021)}} &
\includegraphics[width=0.212\textwidth]{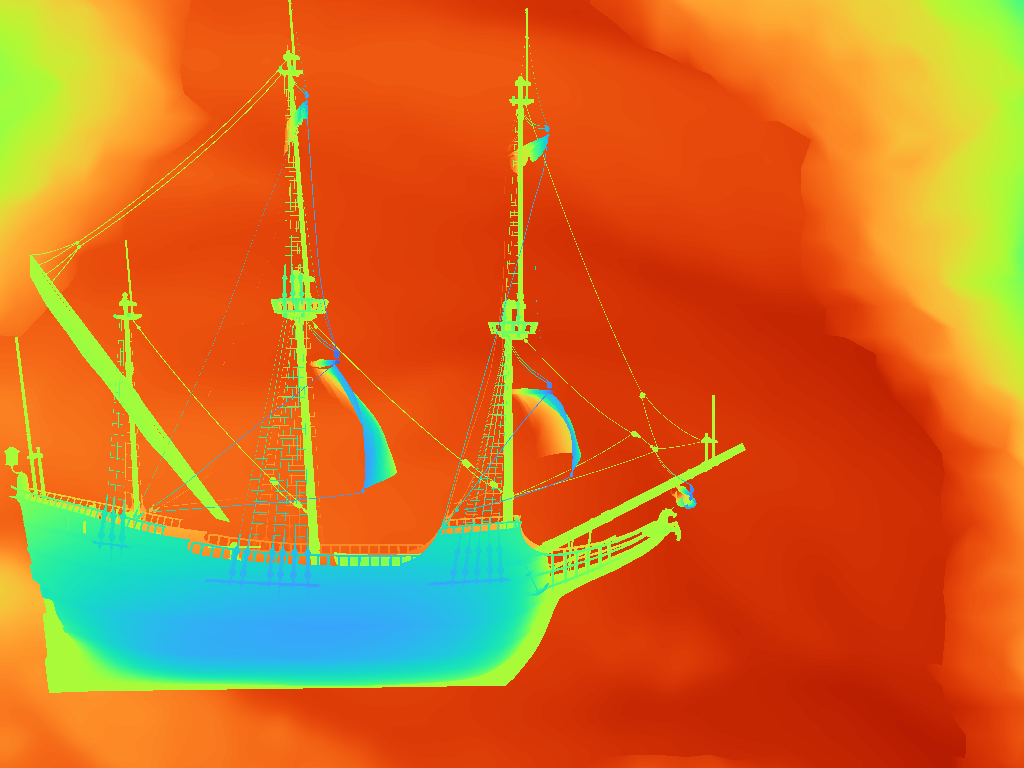} &
\includegraphics[width=0.212\textwidth]{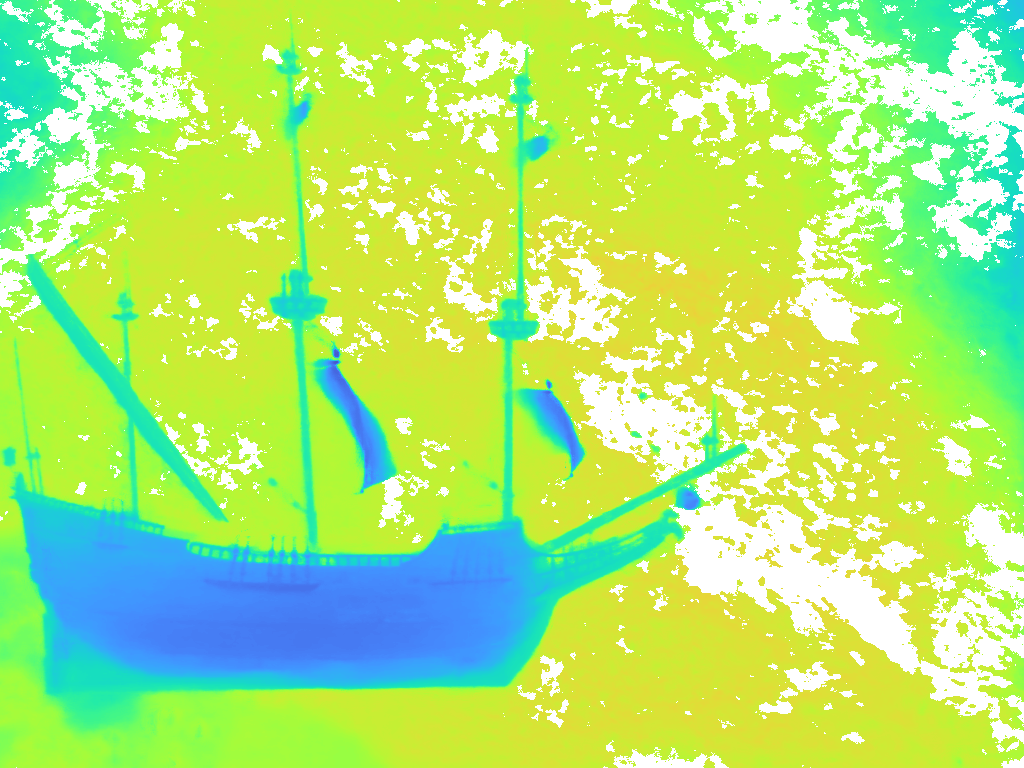} &
\includegraphics[width=0.212\textwidth]{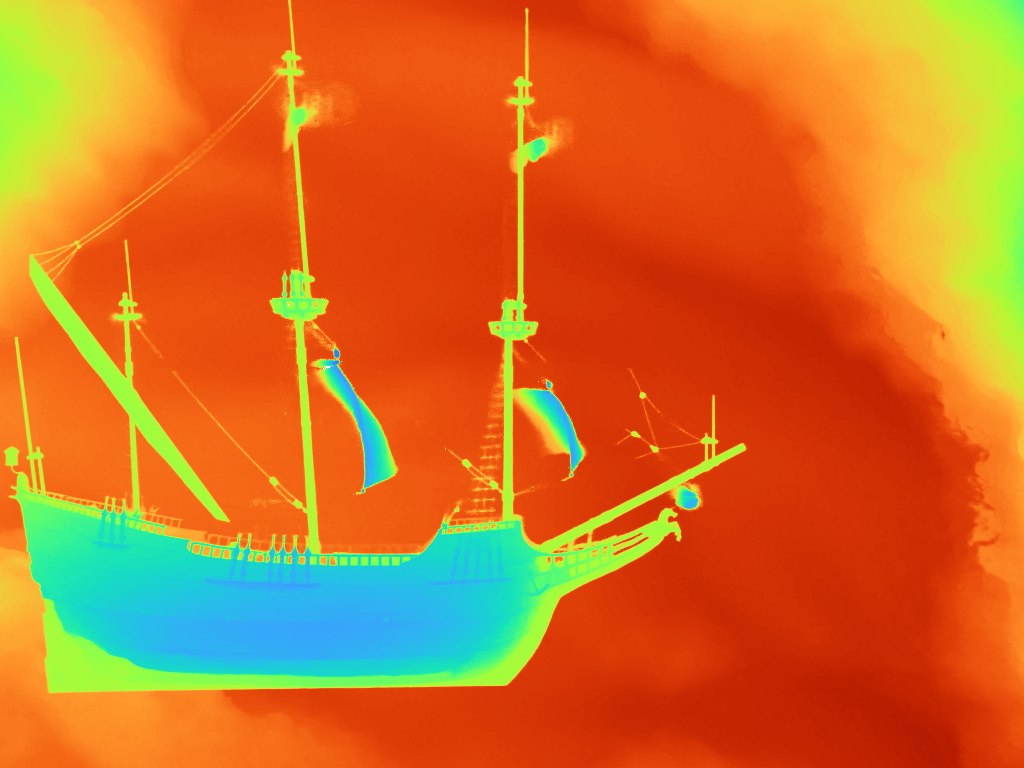} &
\includegraphics[width=0.212\textwidth]{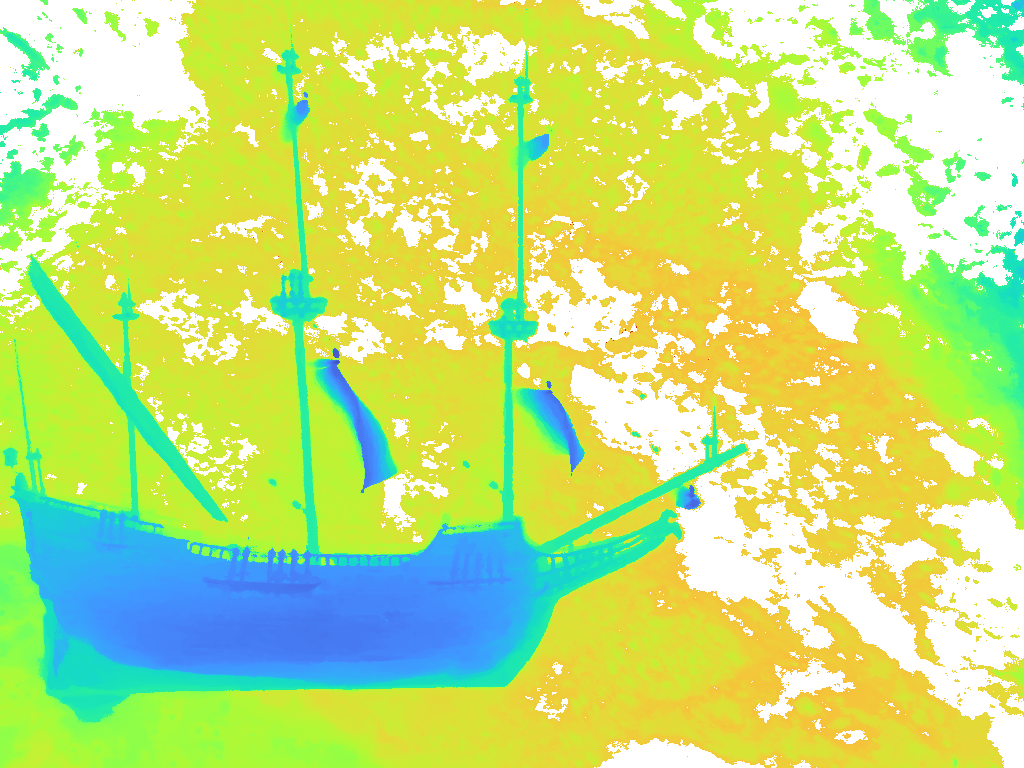} \\[6pt]
\rotatebox{90}{\small\textbf{RGB\,(0080)}} &
\includegraphics[width=0.212\textwidth]{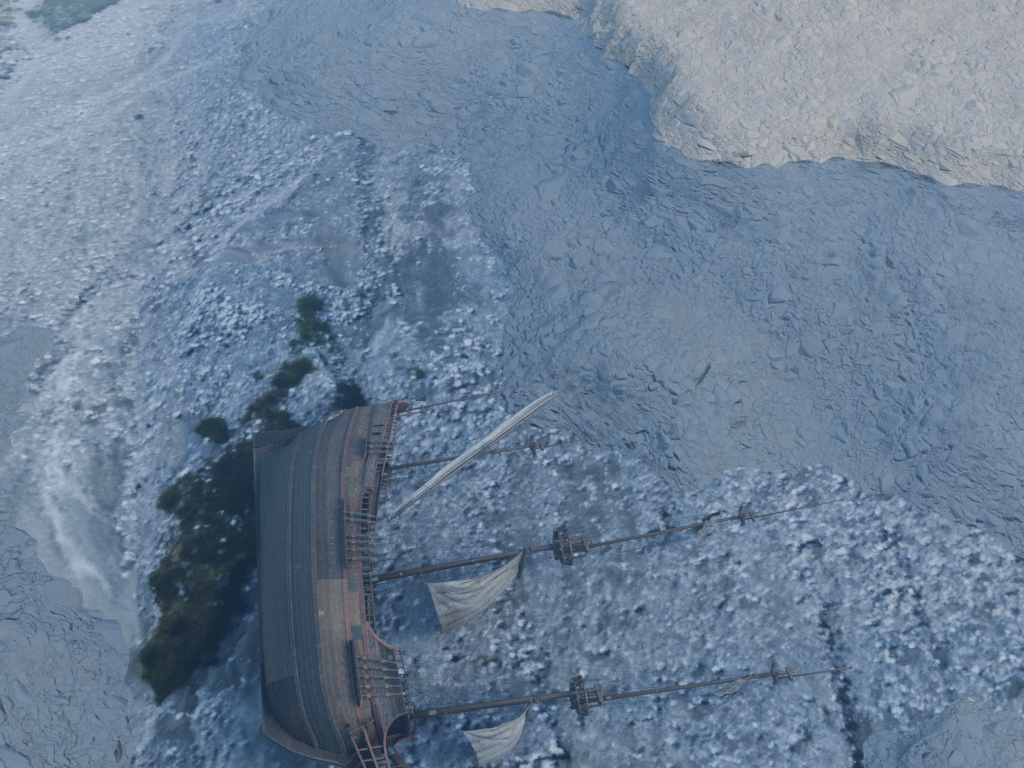} &
\includegraphics[width=0.212\textwidth]{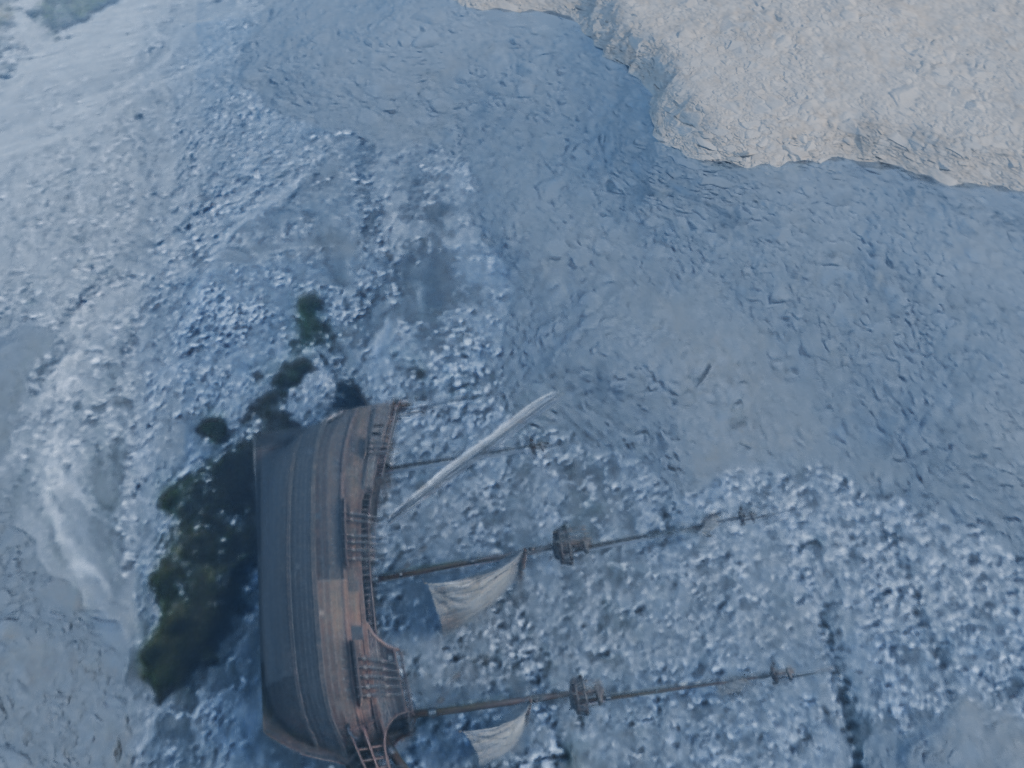} &
\includegraphics[width=0.212\textwidth]{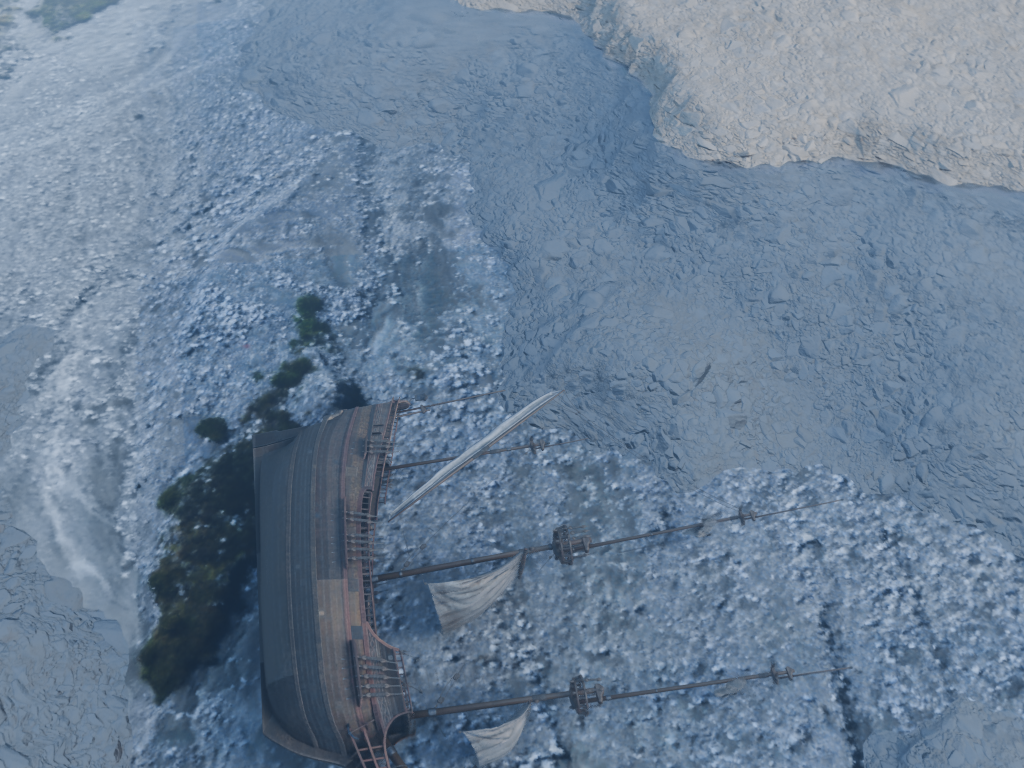} &
\includegraphics[width=0.212\textwidth]{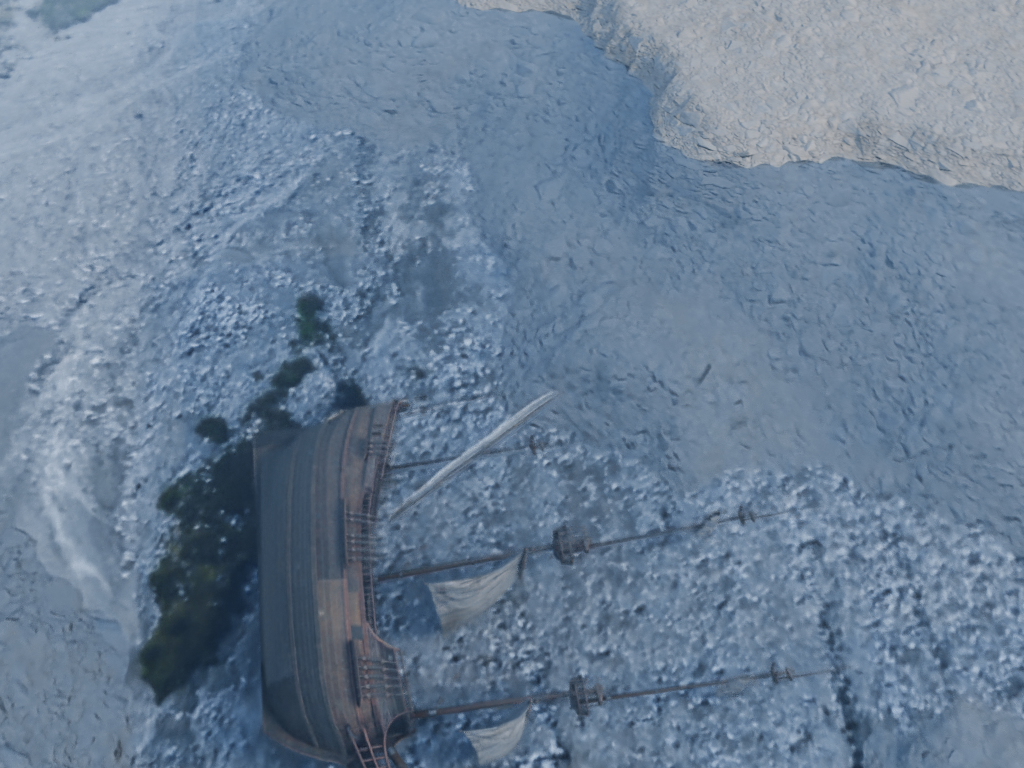} \\[2pt]
\rotatebox{90}{\small\textbf{Depth\,(0080)}} &
\includegraphics[width=0.212\textwidth]{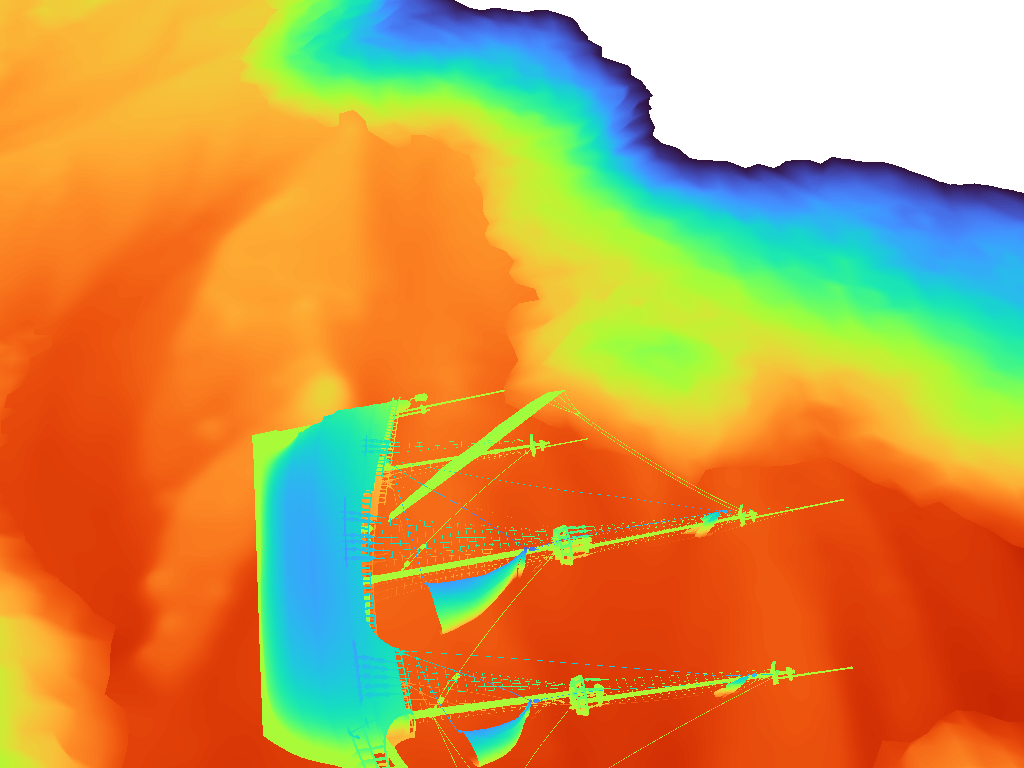} &
\includegraphics[width=0.212\textwidth]{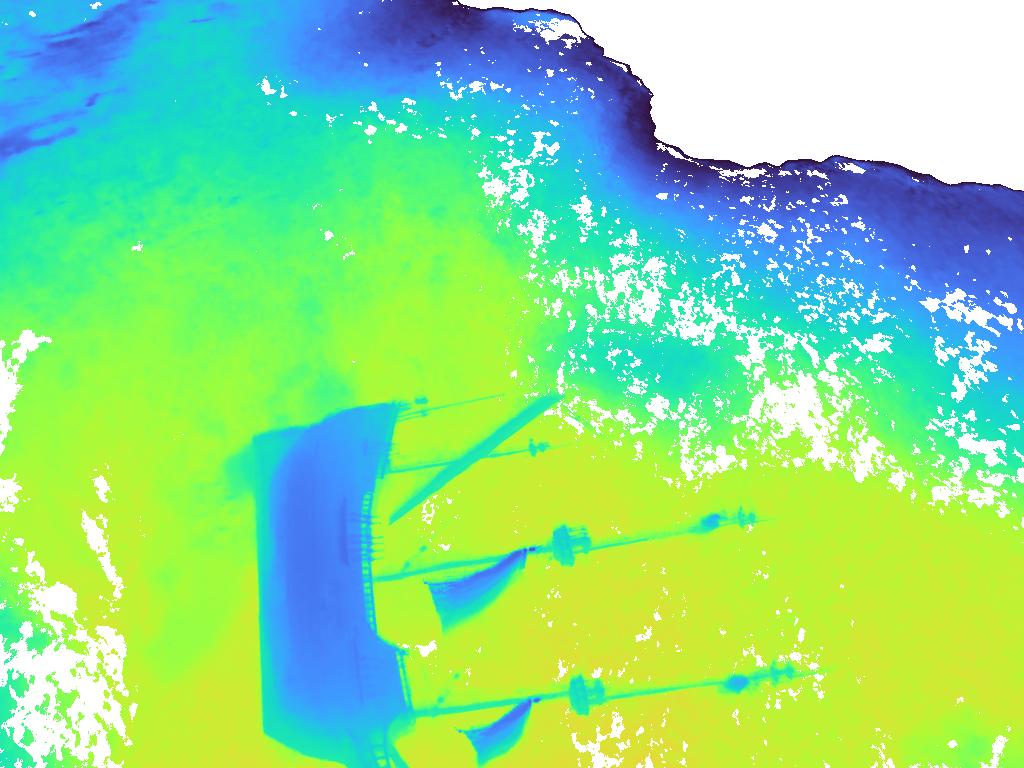} &
\includegraphics[width=0.212\textwidth]{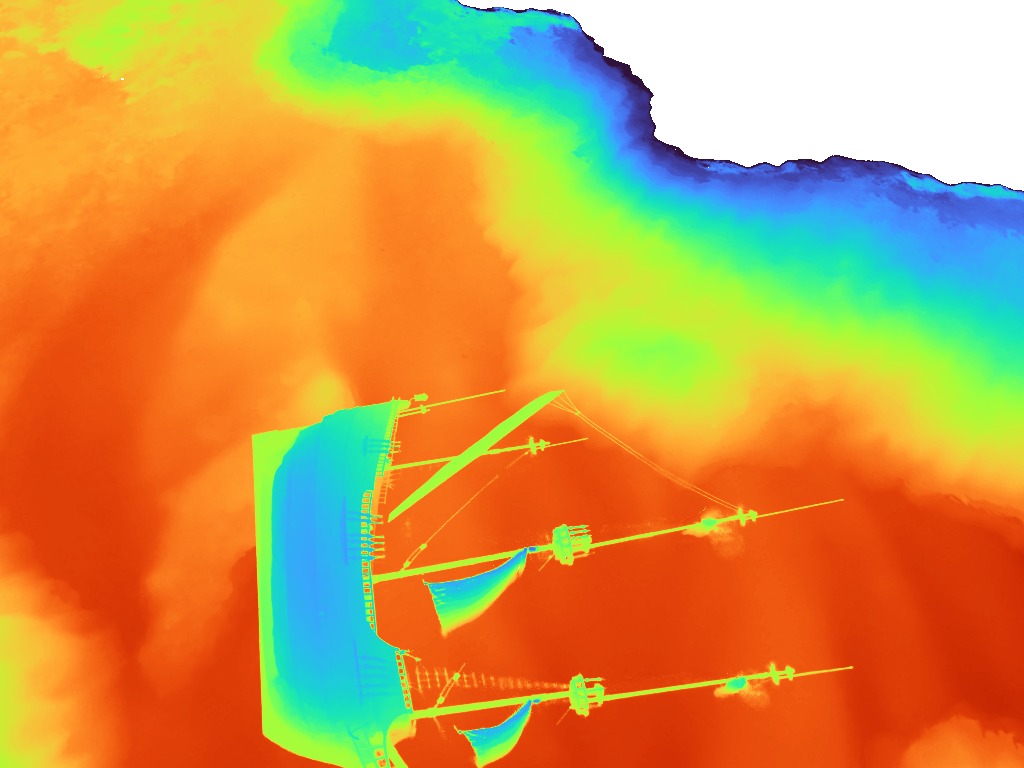} &
\includegraphics[width=0.212\textwidth]{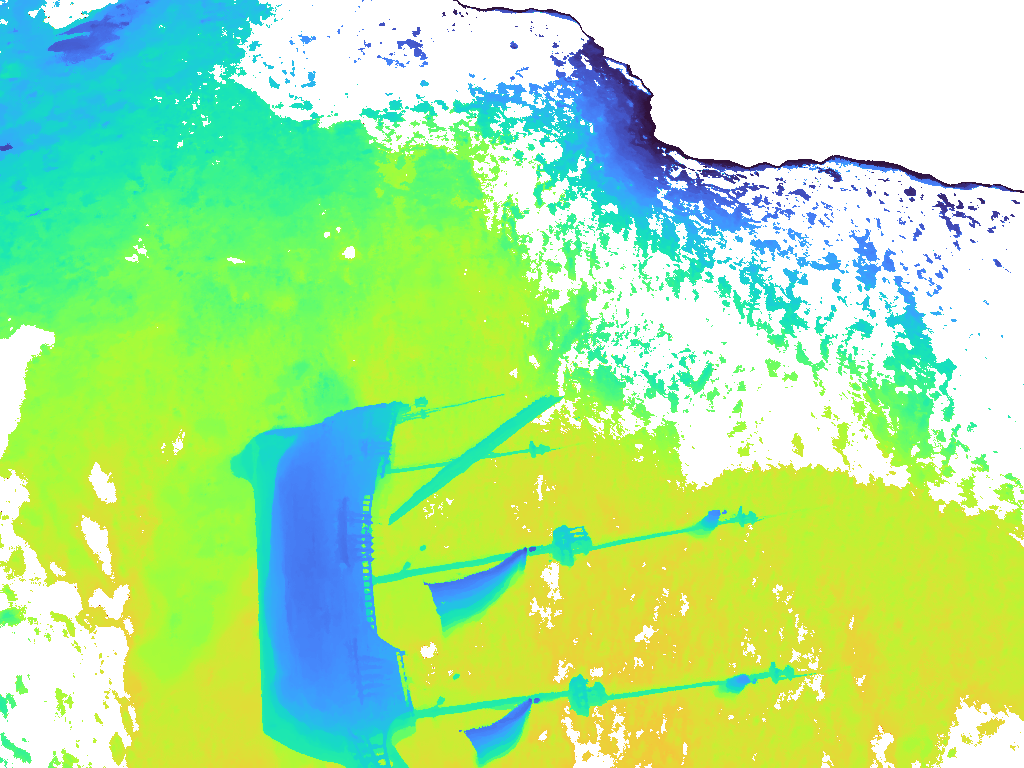} \\[6pt]
\rotatebox{90}{\small\textbf{RGB\,(0004)}} &
\includegraphics[width=0.212\textwidth]{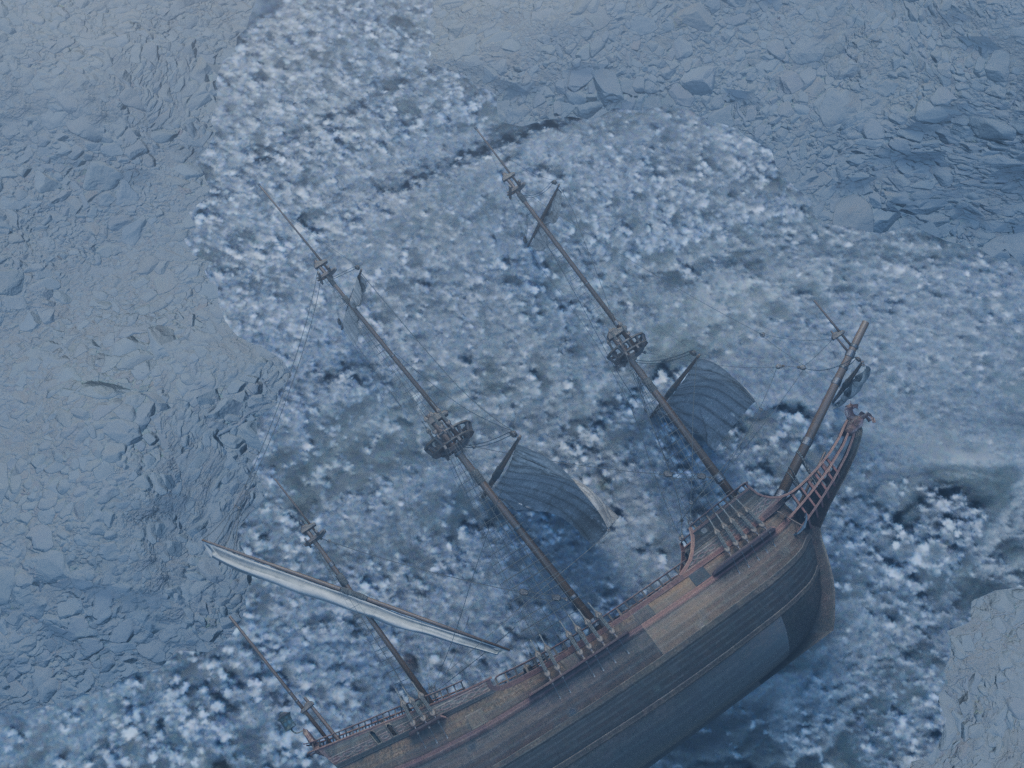} &
\includegraphics[width=0.212\textwidth]{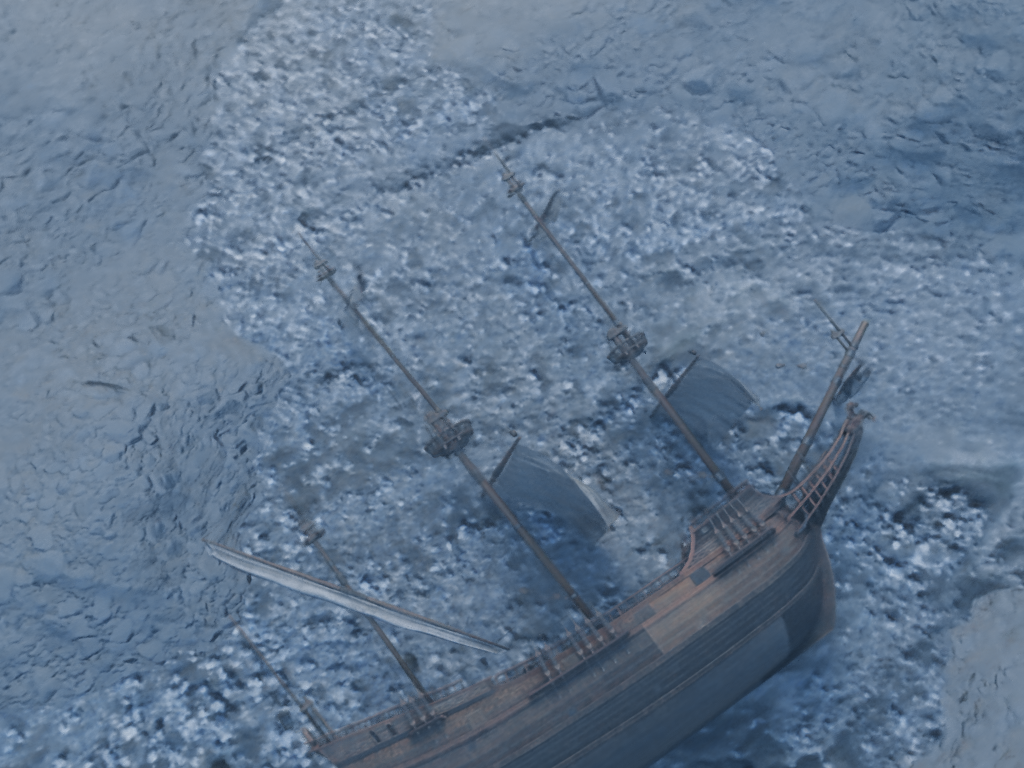} &
\includegraphics[width=0.212\textwidth]{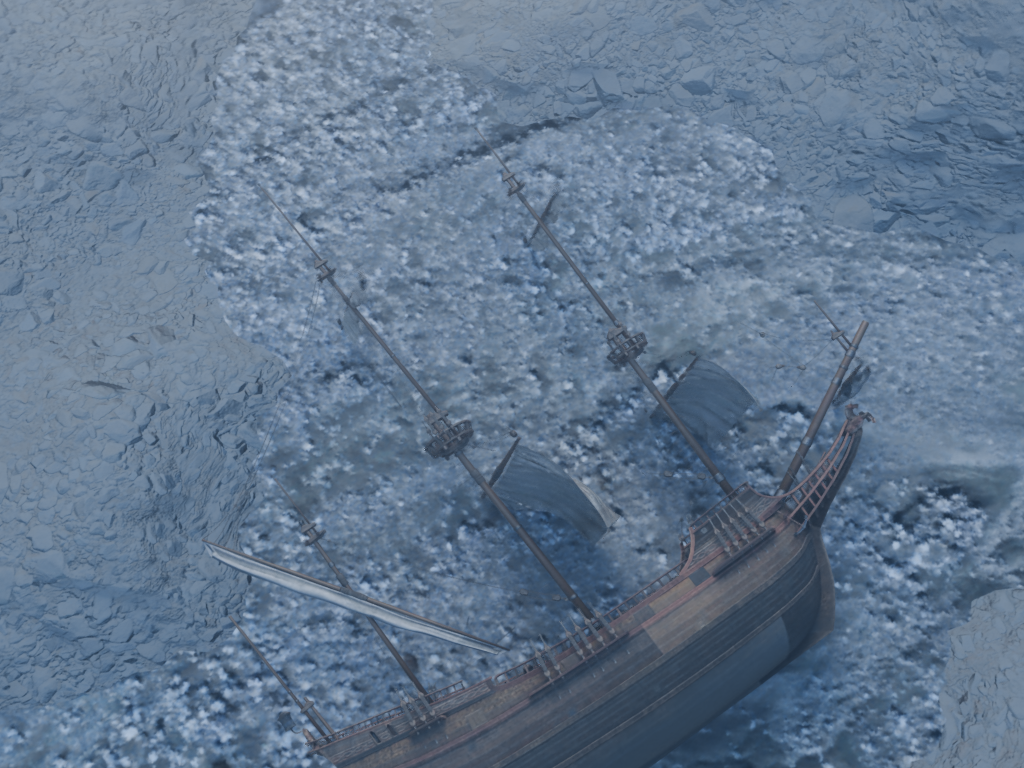} &
\includegraphics[width=0.212\textwidth]{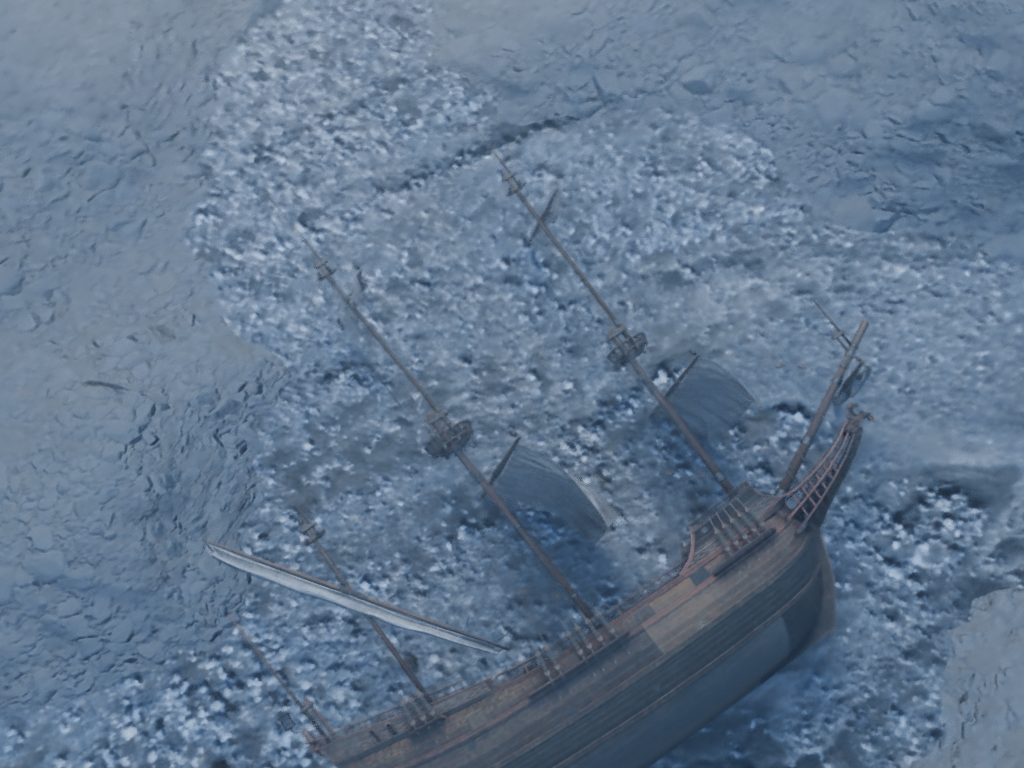} \\[2pt]
\rotatebox{90}{\small\textbf{Depth\,(0004)}} &
\includegraphics[width=0.212\textwidth]{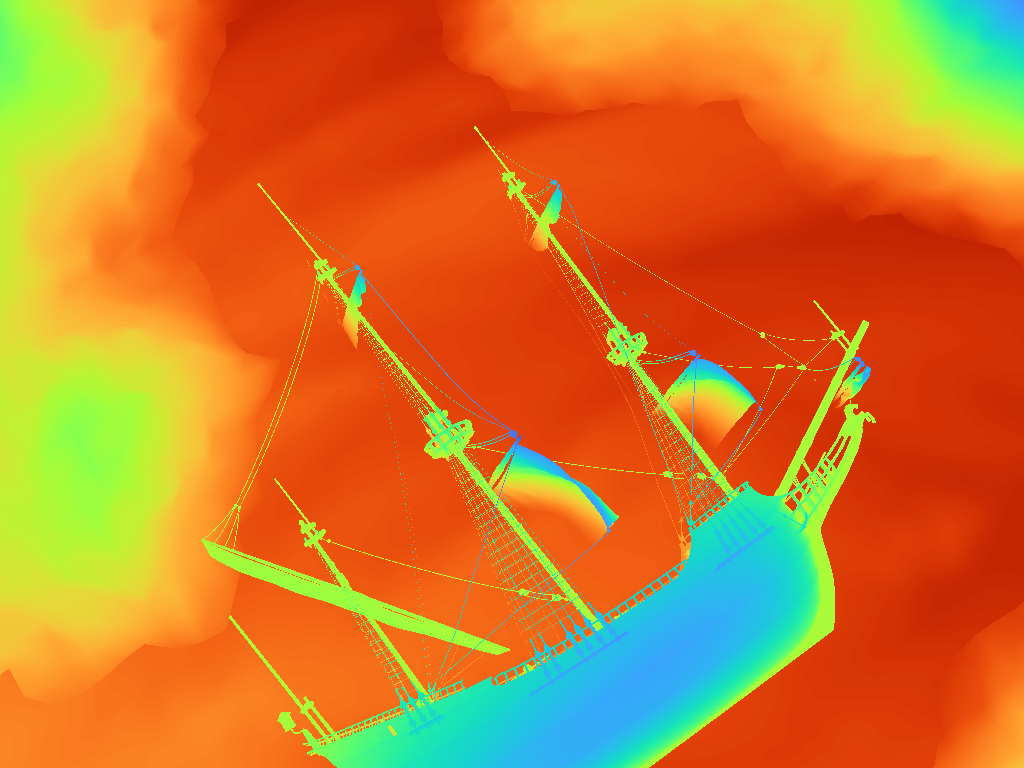} &
\includegraphics[width=0.212\textwidth]{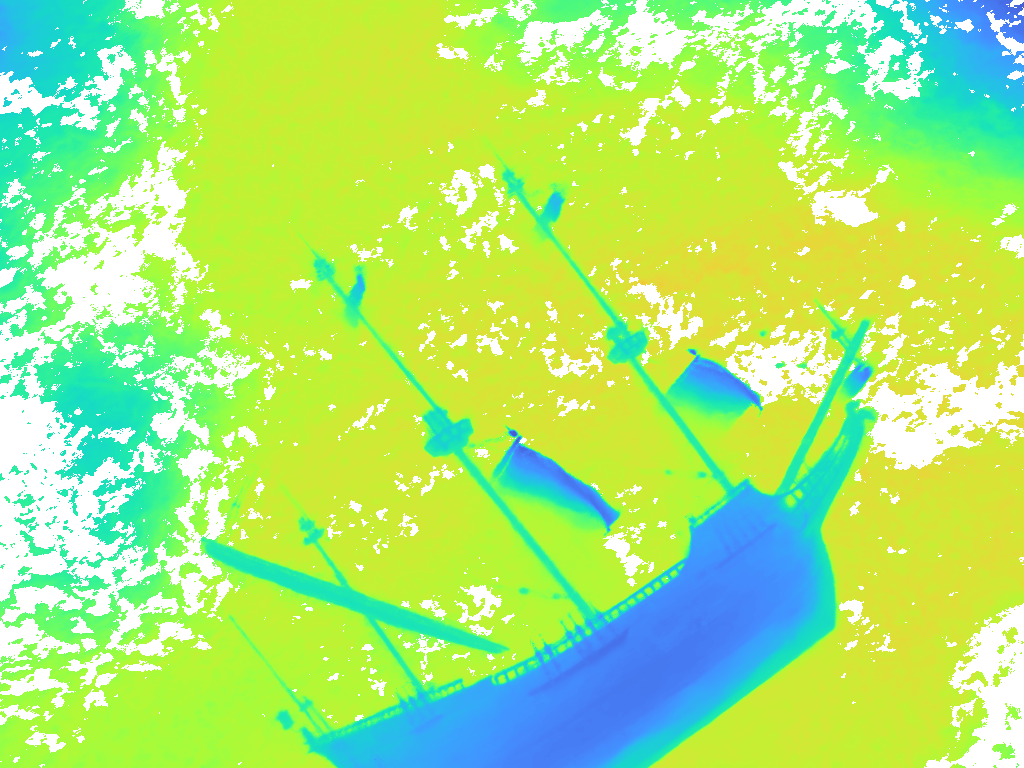} &
\includegraphics[width=0.212\textwidth]{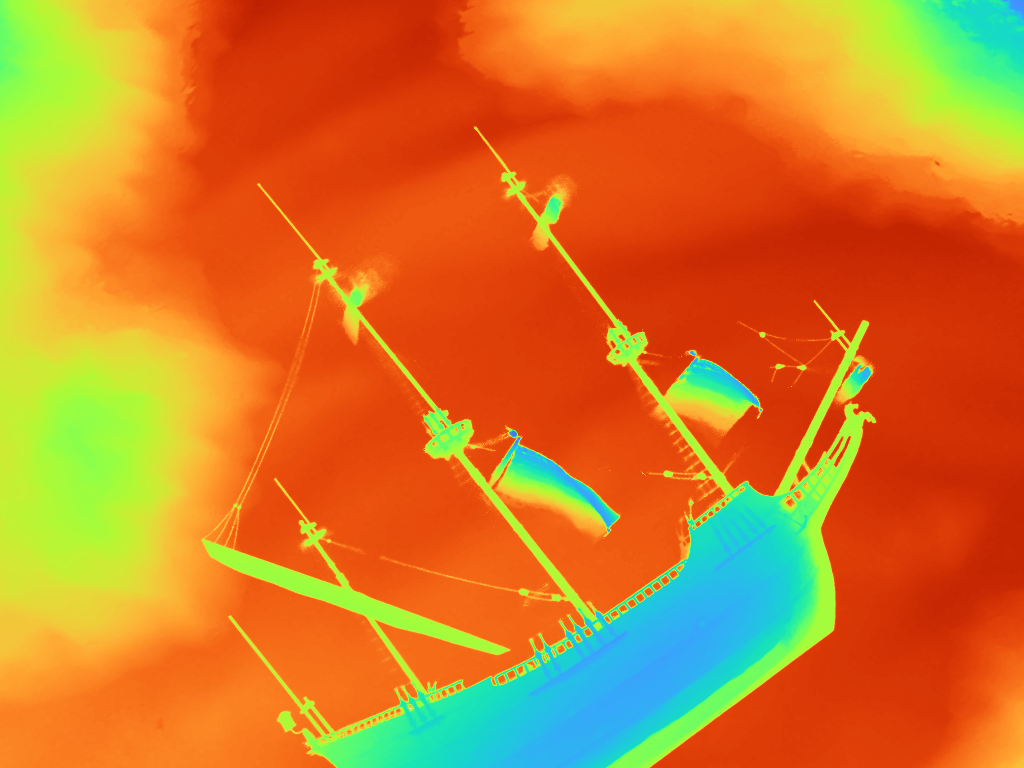} &
\includegraphics[width=0.212\textwidth]{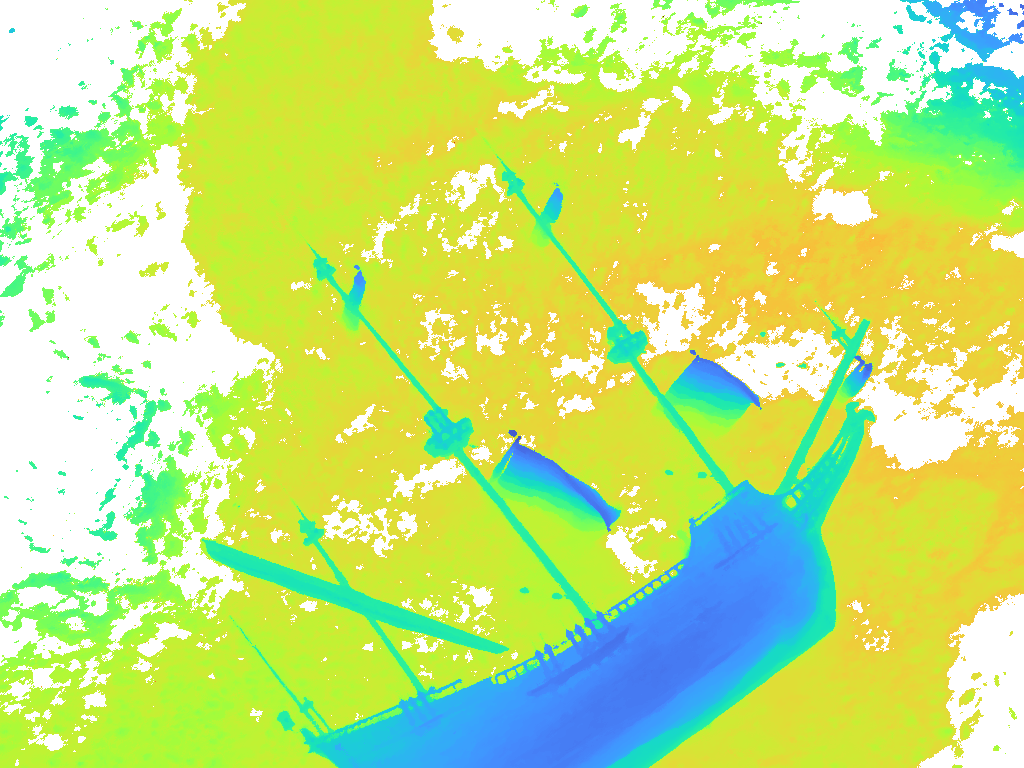} \\
\end{tabular}
\par\smallskip
\includegraphics[width=0.5\textwidth]{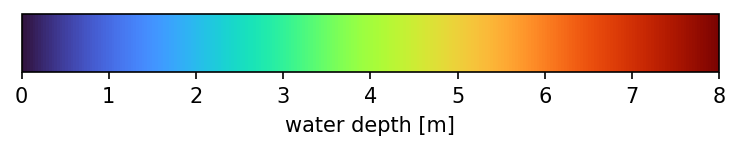}
\caption{Qualitative 2D rendering comparison for three representative evaluation views. Columns: ground truth, BathyFacto without refraction (ablation), BathyFacto with refraction, and Nerfacto baseline. For each view, the top row shows RGB images and the bottom row shows the \emph{water depth below the surface}: the ground-truth depth is obtained by two-segment ray casting against the reference mesh, while the remaining columns show the reconstructed depth of each model. Views span frontal (0021), steep-angle (0080), and oblique (0004) camera geometries. Depth panels are calibrated to metric water depth and use a single shared scale from $0$\,m (water surface) to $8$\,m, identical across all views and methods (see colorbar); cooler colors are shallower (near the surface) and warmer colors are deeper. Non-water pixels (sky, land, and the water surface) are shown in white. The refraction-aware model recovers the seafloor at its true depth (warm), whereas the non-refracted baselines reconstruct it too shallow (cool).
}
\label{fig:qualitative-grid}
\end{figure*}

\autoref{fig:qualitative-grid} shows rendered RGB images and depth maps for three representative evaluation views. All three configurations produce visually plausible RGB renderings. The depth maps, however, reveal clear differences: BathyFacto with refraction produces smooth, spatially consistent depth predictions across the underwater region. The no-refraction ablation shows a visible depth discontinuity at the water surface, where the straight-ray model fails to account for the change in optical path length. The Nerfacto baseline exhibits floater artifacts on the seabed, visible as scattered depth outliers in the depth maps, indicating incomplete density convergence in the water column.

\subsection{Quantitative Results}
Table~\ref{tab:quantitative-results} reports the 2D image-space evaluation for the Nerfacto baseline, BathyFacto with refraction, and the BathyFacto no-refraction ablation.

\textbf{Evaluation protocol.}
All methods are evaluated on the same \texttt{eval} split (\(13\) images). For each image, predictions are computed and valid-pixel masks are applied consistently across all models to remove ``no data''-pixels. PSNR, SSIM, and LPIPS are computed on the masked images and averaged over all evaluation views.

\begin{table}[t]
\centering
\caption{2D image-space evaluation on the simulation \texttt{eval} split (mean over \(13\) images). Metrics are computed over all valid pixels. In contrast to the 3D evaluation (Table~\ref{tab:quantitative-3d-results}), which is restricted to the central crop, these 2D metrics are computed over the full evaluation images.}
\label{tab:quantitative-results}
\small
\setlength{\tabcolsep}{3pt}
\renewcommand{\arraystretch}{1.1}
\begin{tabularx}{\columnwidth}{@{}
    >{\raggedright\arraybackslash}X
    >{\centering\arraybackslash}p{0.15\columnwidth}
    >{\centering\arraybackslash}p{0.15\columnwidth}
    >{\centering\arraybackslash}p{0.15\columnwidth}@{}}
\toprule
\textbf{Method} & \textbf{PSNR} $\uparrow$ & \textbf{SSIM} $\uparrow$ & $\textbf{LPIPS} \ \smash{\downarrow}$ \\
\midrule
Nerfacto                 & 25.72          & 0.561          & 0.256 \\
BathyFacto (Refr.\ OFF) & 32.43          & 0.790          & 0.180 \\
BathyFacto (Refr.\ ON)  & \textbf{34.18} & \textbf{0.893} & \textbf{0.095} \\
\bottomrule
\end{tabularx}
\end{table}

BathyFacto with refraction achieves the highest PSNR (34.18\,dB), SSIM (0.893), and best LPIPS (0.095), followed by BathyFacto without refraction (32.43\,dB PSNR). These results are consistent with previous evaluations of two-media NeRFs \citep{gunthner2025_NeRFracApplicationToUAV,brezovsky2025_Analysis} and demonstrate that the accuracy of the 3D geometry is the decisive quality criterion for bathymetric reconstruction, rather than image-space metrics.

\noindent\textbf{Interpretation note.}
As discussed in Section~\ref{sec:experimental-setup}, 2D image-space metrics do not directly imply superior 3D geometric accuracy; a method may achieve high PSNR by fitting pixel colors with geometrically incorrect depth. The 3D point-cloud evaluation in the following section therefore remains the decisive criterion for geometric validation.

\subsection{3D Point-Cloud Evaluation}
\begin{figure*}[p]
    \centering

    \small\textbf{BathyFacto (Refr.\ ON)}\\[2pt]
    \includegraphics[width=0.64\textwidth]{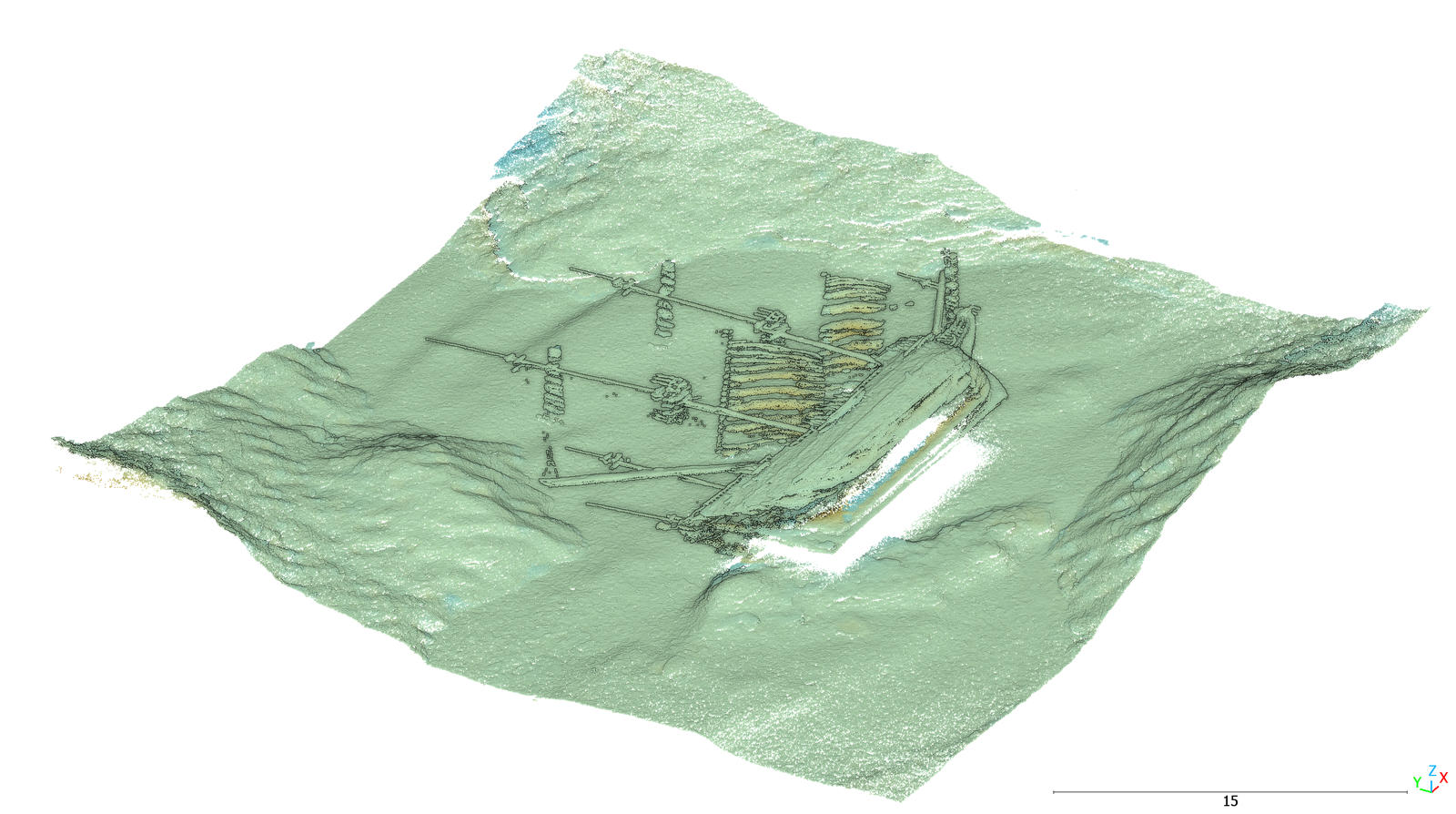}

    \vspace{4pt}
    \setlength{\tabcolsep}{3pt}
    \begin{tabular}{@{}cc@{}}
        \small\textbf{Nerfacto (uncorrected)} & \small\textbf{Nerfacto ($+1/3$ depth)} \\[2pt]
        \includegraphics[width=0.38\textwidth]{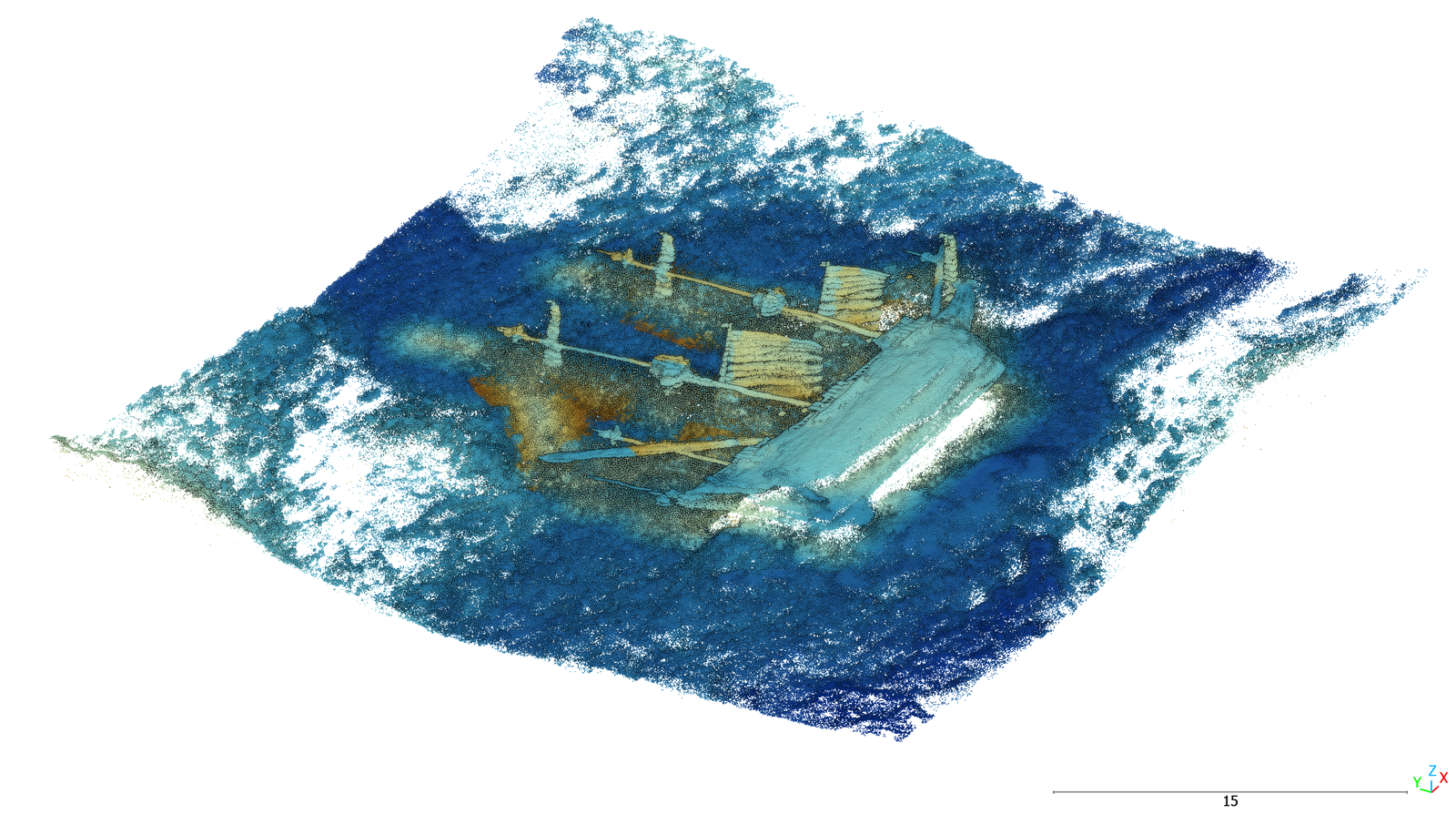} &
        \includegraphics[width=0.38\textwidth]{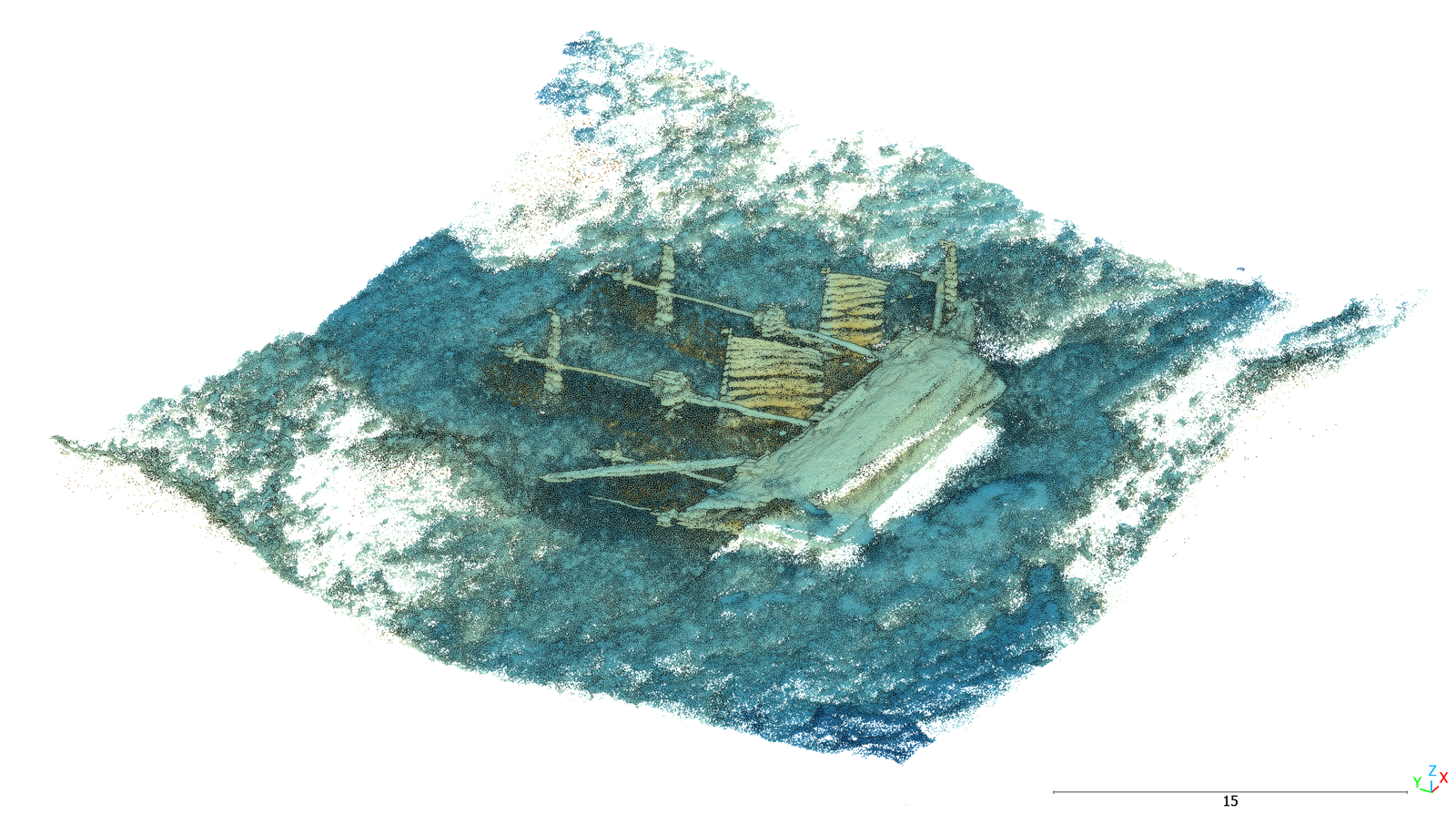} \\[4pt]
        \small\textbf{BathyFacto Refr.\ OFF (uncorrected)} & \small\textbf{BathyFacto Refr.\ OFF ($+1/3$ depth)} \\[2pt]
        \includegraphics[width=0.38\textwidth]{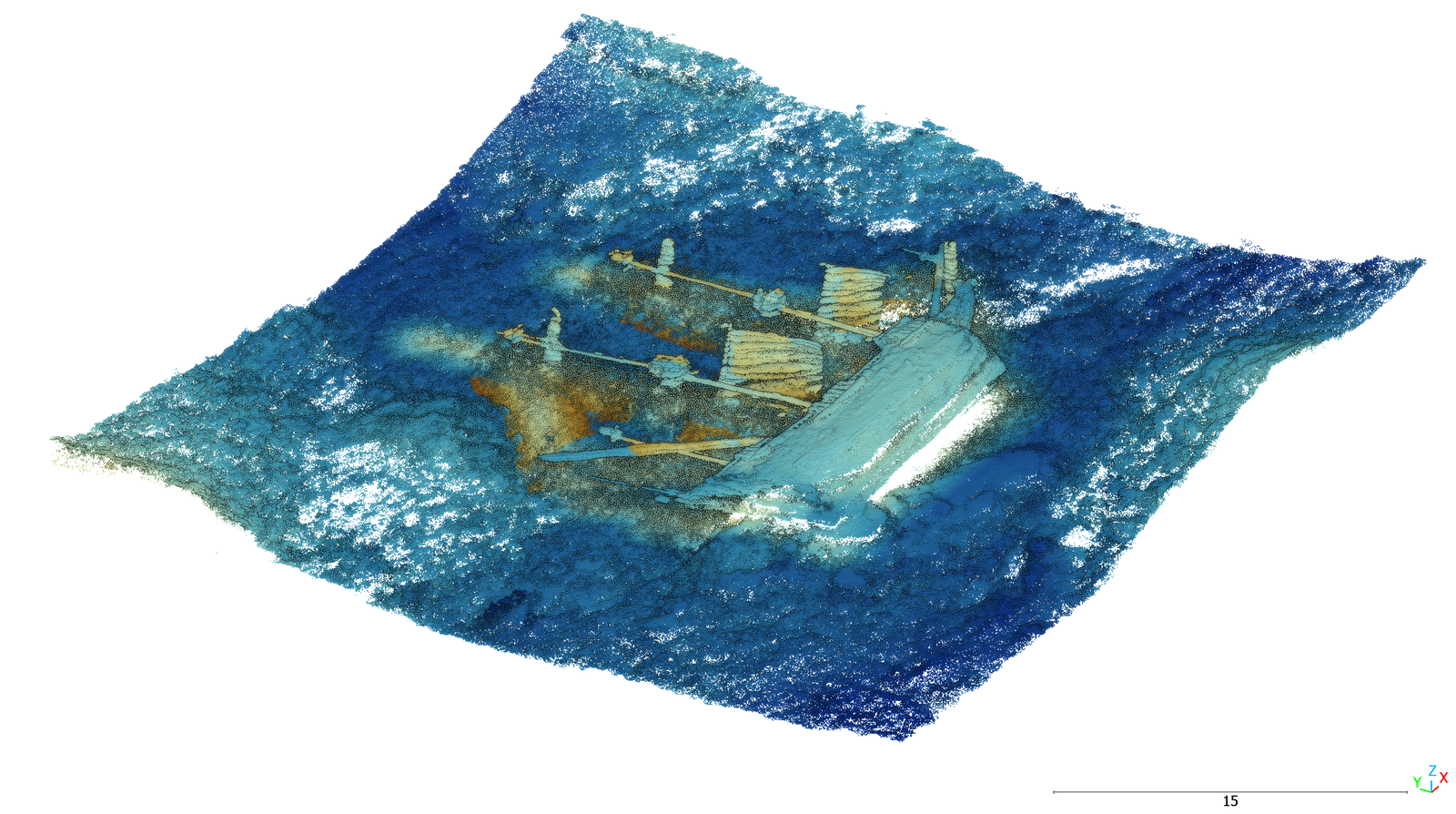} &
        \includegraphics[width=0.38\textwidth]{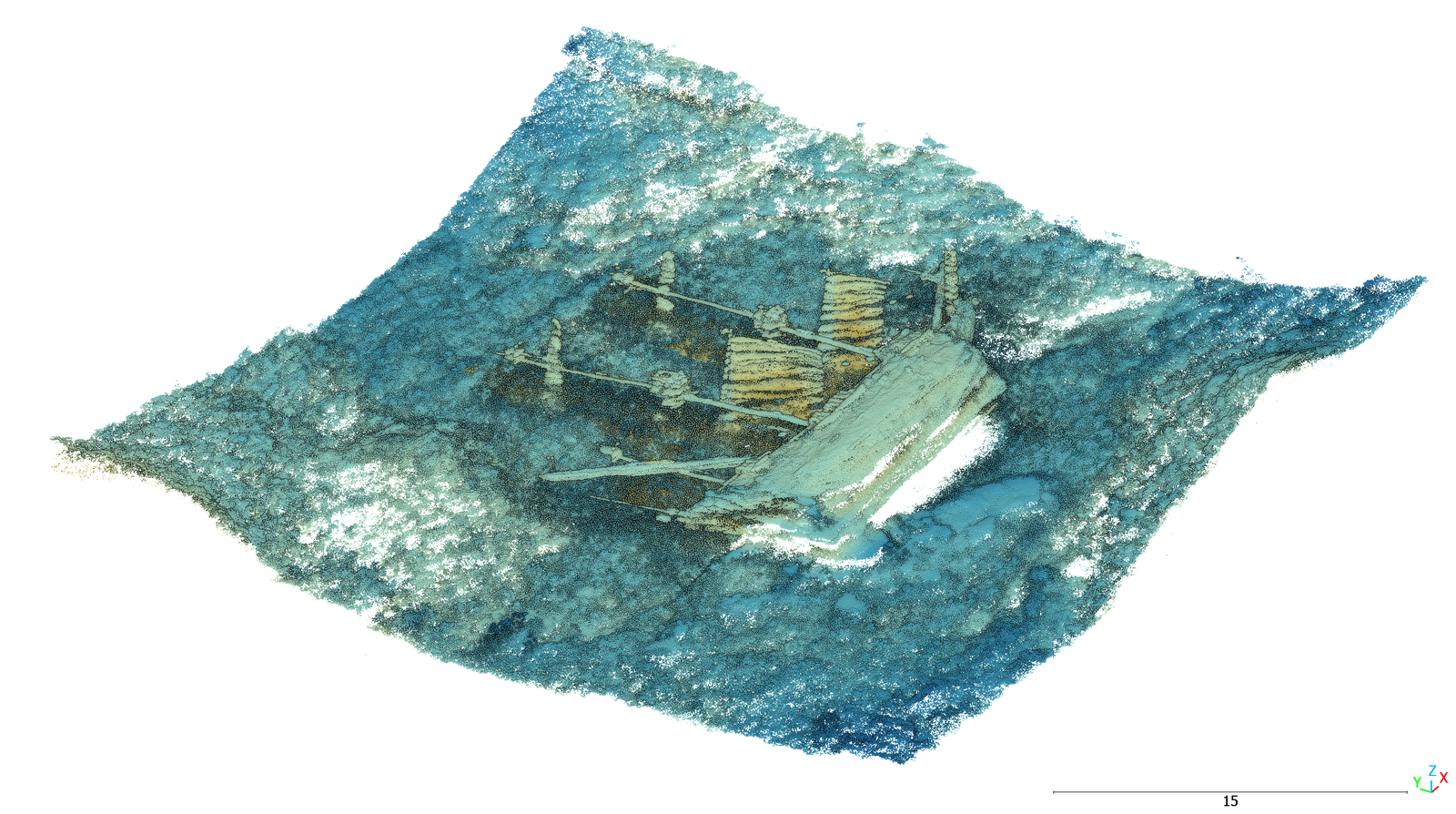} \\[4pt]
        \small\textbf{MVS (uncorrected, apparent depth)} & \small\textbf{MVS (refr.-corrected)} \\[2pt]
        \includegraphics[width=0.38\textwidth]{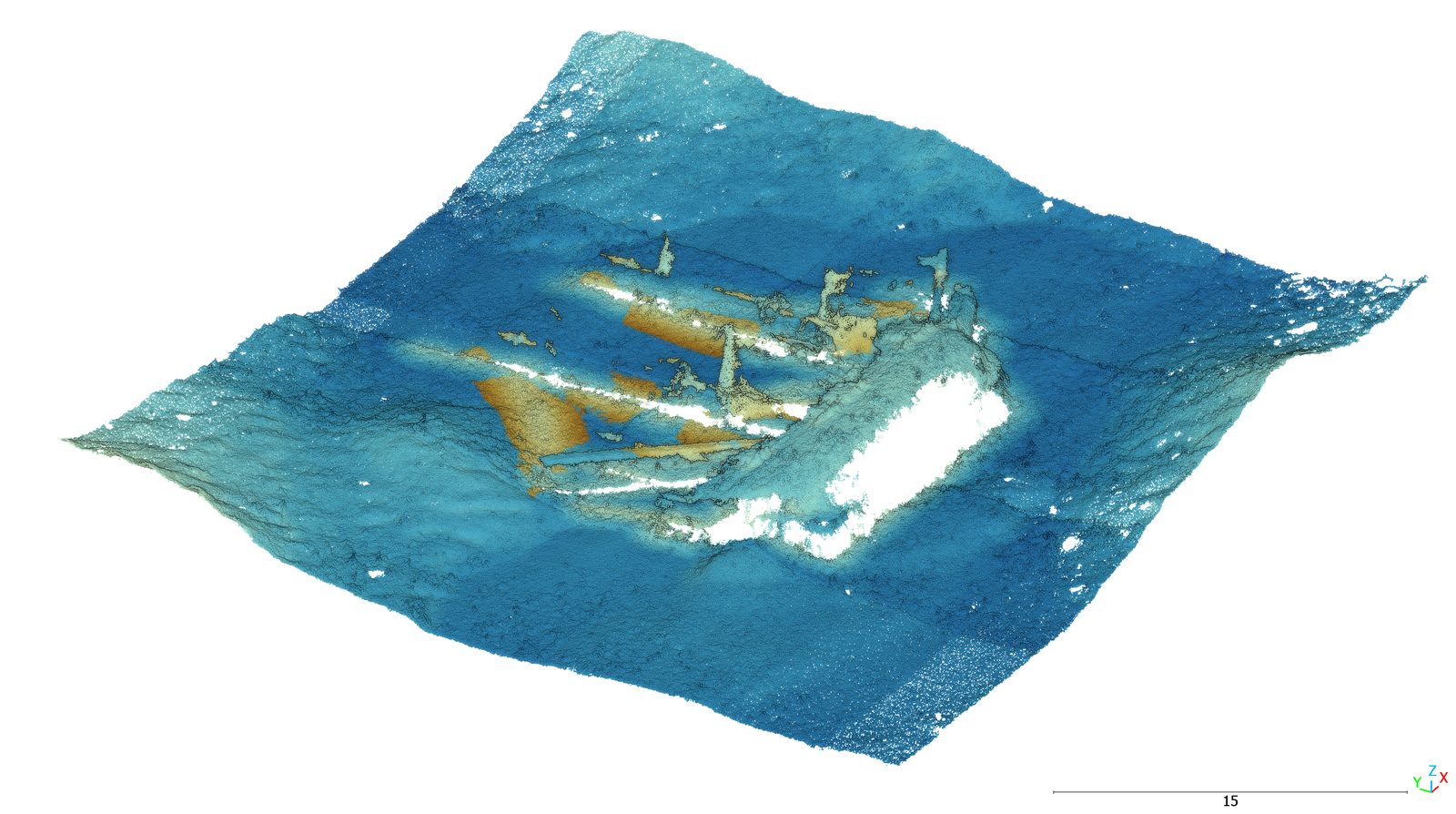} &
        \includegraphics[width=0.38\textwidth]{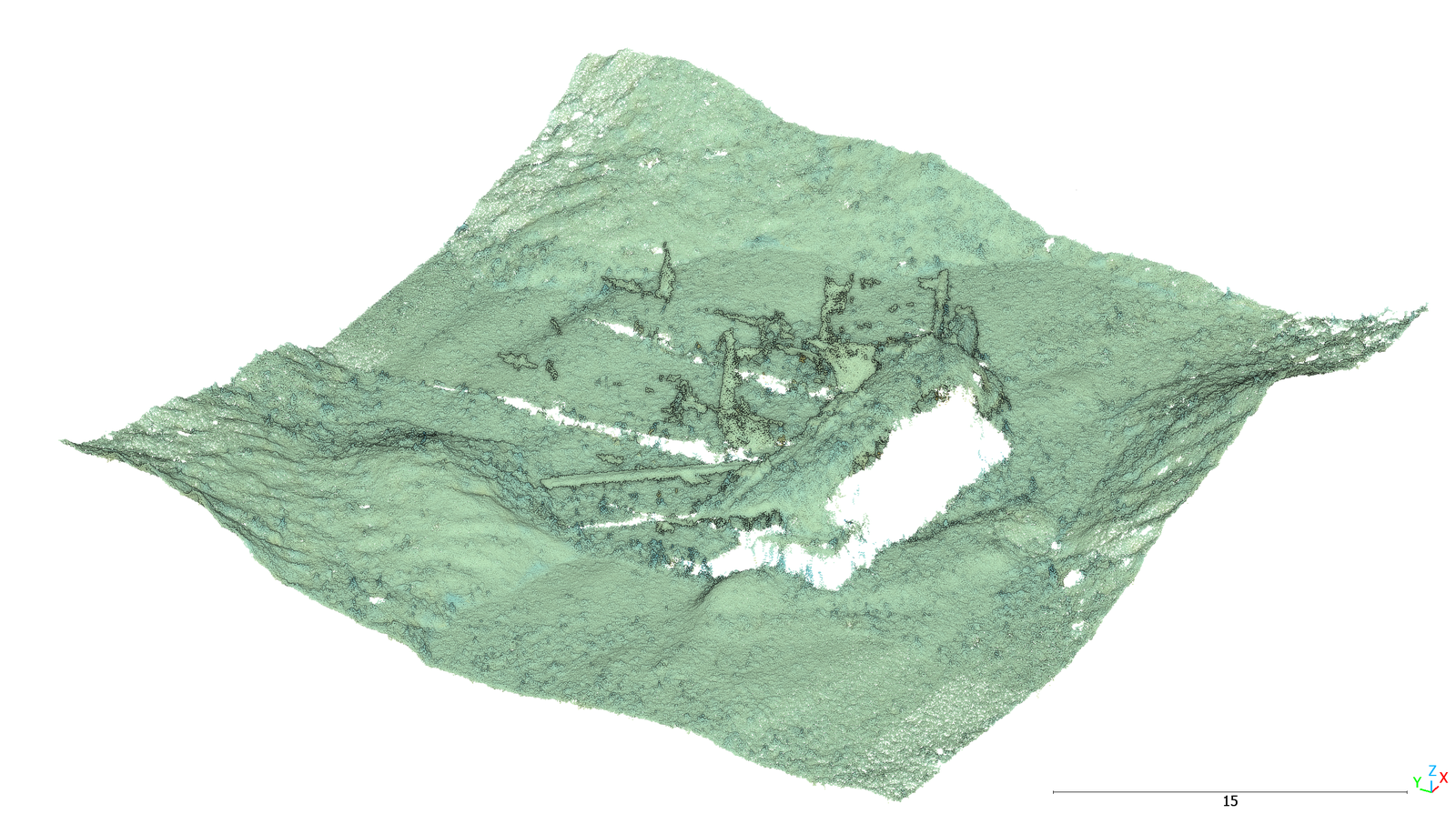} \\
    \end{tabular}

    \vspace{2pt}
    \includegraphics[width=0.65\textwidth]{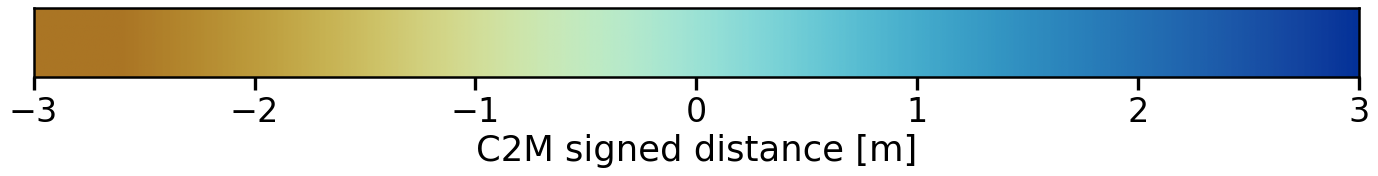}

    \caption{Cloud-to-Mesh (C2M) signed distance maps. The matrix mirrors Table~\ref{tab:quantitative-3d-results}: the left column shows each method uncorrected, the right column its corrected variant, with the correction named per panel --- the $+1/3$-depth correction is the approximate correction for the non-refracted NeRF baselines, while the per-ray Snell correction of the MVS cloud (Sect.~\ref{sec:experimental-setup:configs}) makes it a genuine refraction-aware baseline. BathyFacto (Refr.\ ON) needs no correction and is shown as-is in the absolute global frame (top); it achieves a C2M signed median of $-0.001$\,m and 85.7\,\% completeness at 0.2\,m tolerance. All panels share one fixed color scale ($-3$\,\ldots\,$+3$\,m).}
    \label{fig:3d-evaluation-page}
\end{figure*}

\begin{figure*}[p]
    \centering

    \textbf{(a) C2M signed distance histograms}\par\smallskip
    \setlength{\tabcolsep}{3pt}
    \begin{tabular}{@{}c@{\hspace{1pt}}cc@{\hspace{1pt}}c@{}}
        & \small\textbf{Nerfacto ($+1/3$ depth)} & & \small\textbf{BathyFacto Refr.\ OFF ($+1/3$ depth)} \\[3pt]
        \adjustbox{valign=m}{\rotatebox{90}{\small Count}} &
        \includegraphics[width=0.43\textwidth,valign=m]{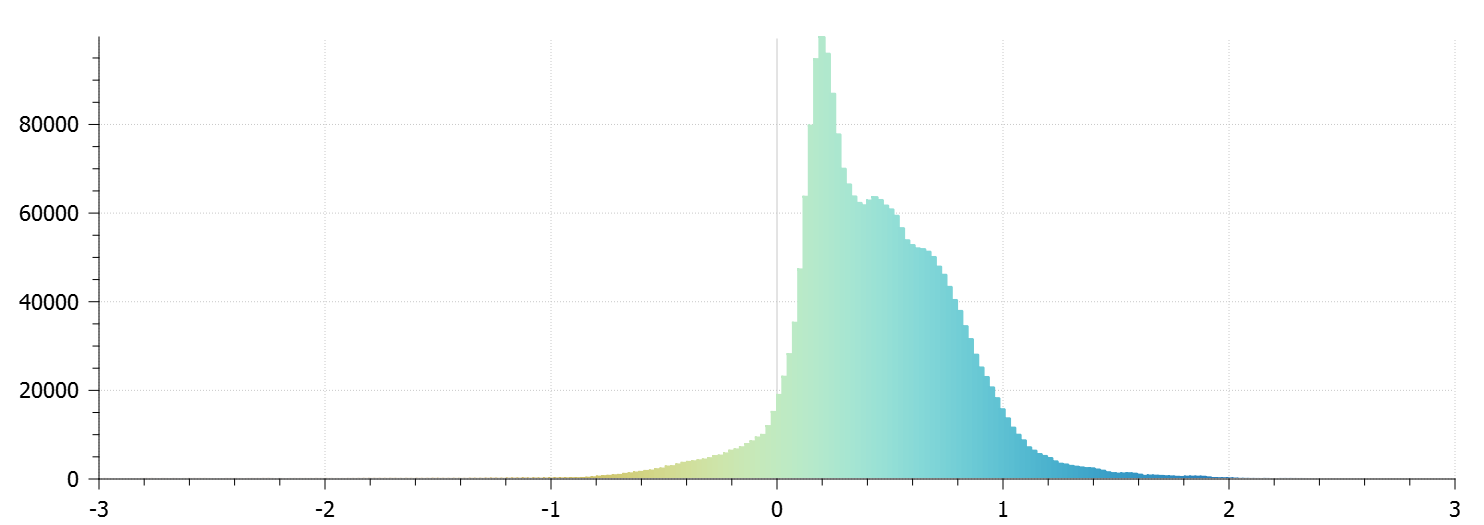} &
        \adjustbox{valign=m}{\rotatebox{90}{\small Count}} &
        \includegraphics[width=0.43\textwidth,valign=m]{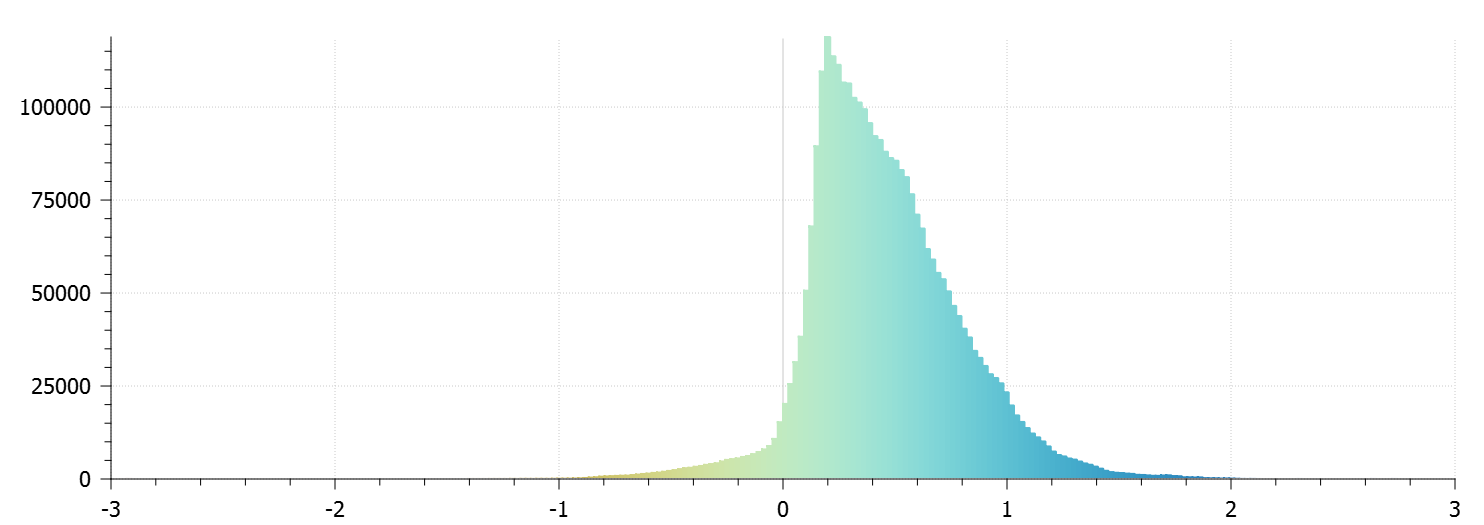} \\
        & \small C2M signed distance [m] & & \small C2M signed distance [m] \\[6pt]
        & \small\textbf{BathyFacto (Refr.\ ON)} & & \small\textbf{MVS (refr.-corrected)} \\[3pt]
        \adjustbox{valign=m}{\rotatebox{90}{\small Count}} &
        \includegraphics[width=0.43\textwidth,valign=m]{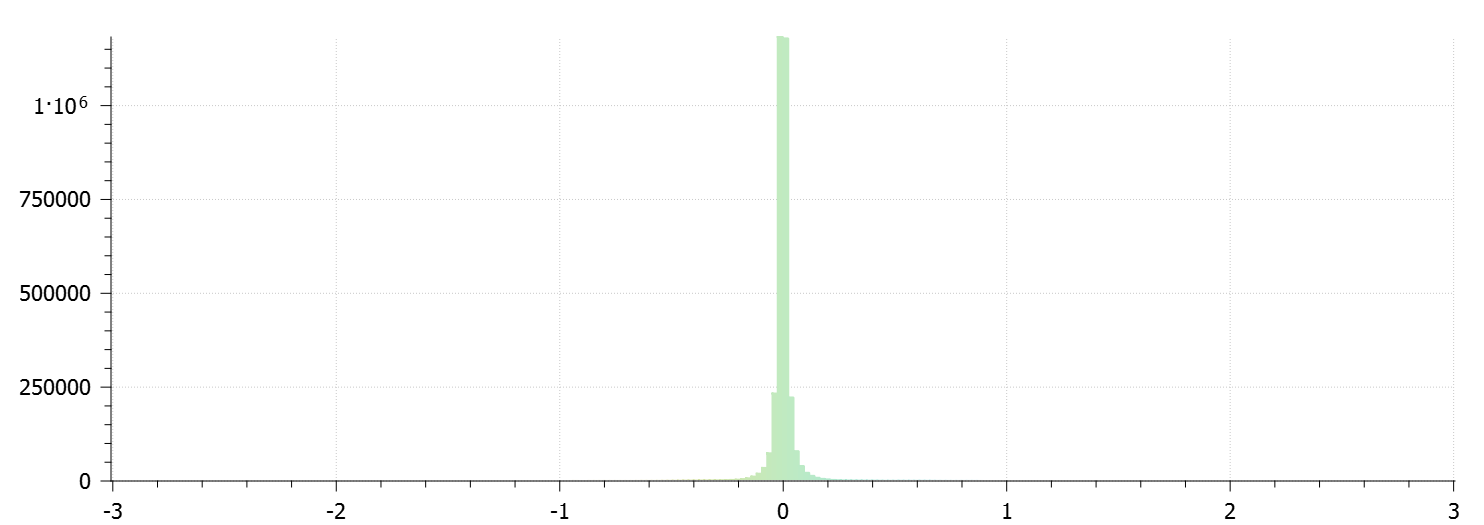} &
        \adjustbox{valign=m}{\rotatebox{90}{\small Count}} &
        \includegraphics[width=0.43\textwidth,valign=m]{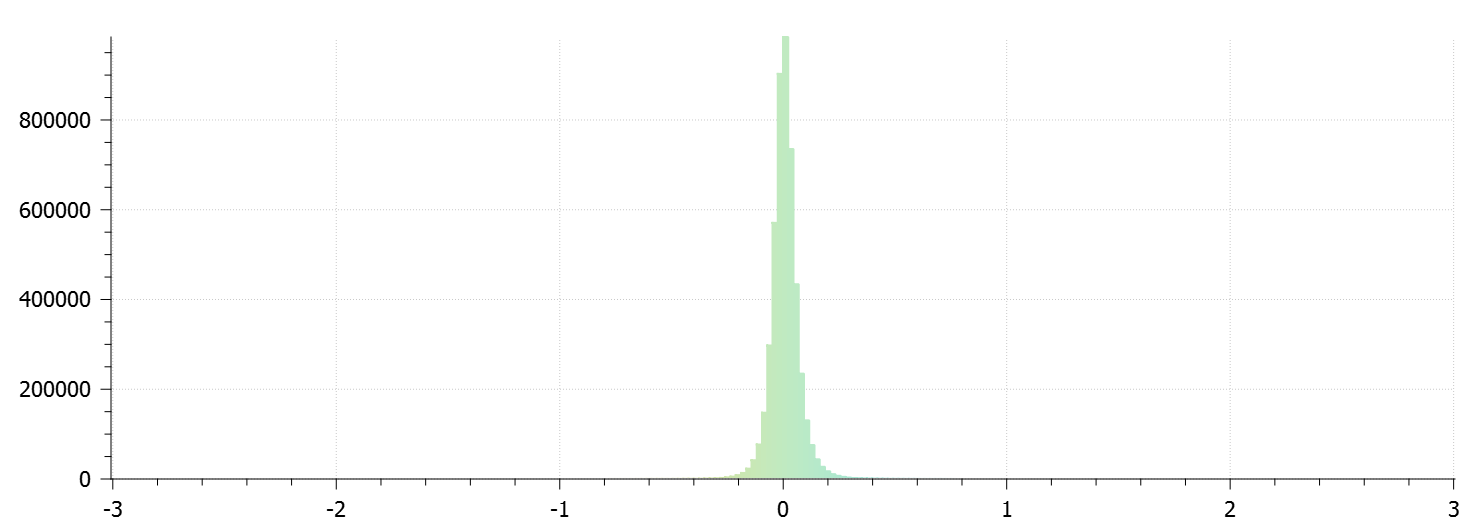} \\
        & \small C2M signed distance [m] & & \small C2M signed distance [m] \\
    \end{tabular}

    \vspace{6pt}

    \textbf{(b) Completeness maps}\par\smallskip
    \setlength{\tabcolsep}{3pt}
    \begin{tabular}{@{}cc@{}}
        \small\textbf{Nerfacto ($+1/3$ depth)} & \small\textbf{BathyFacto Refr.\ OFF ($+1/3$ depth)} \\[3pt]
        \includegraphics[width=0.43\textwidth]{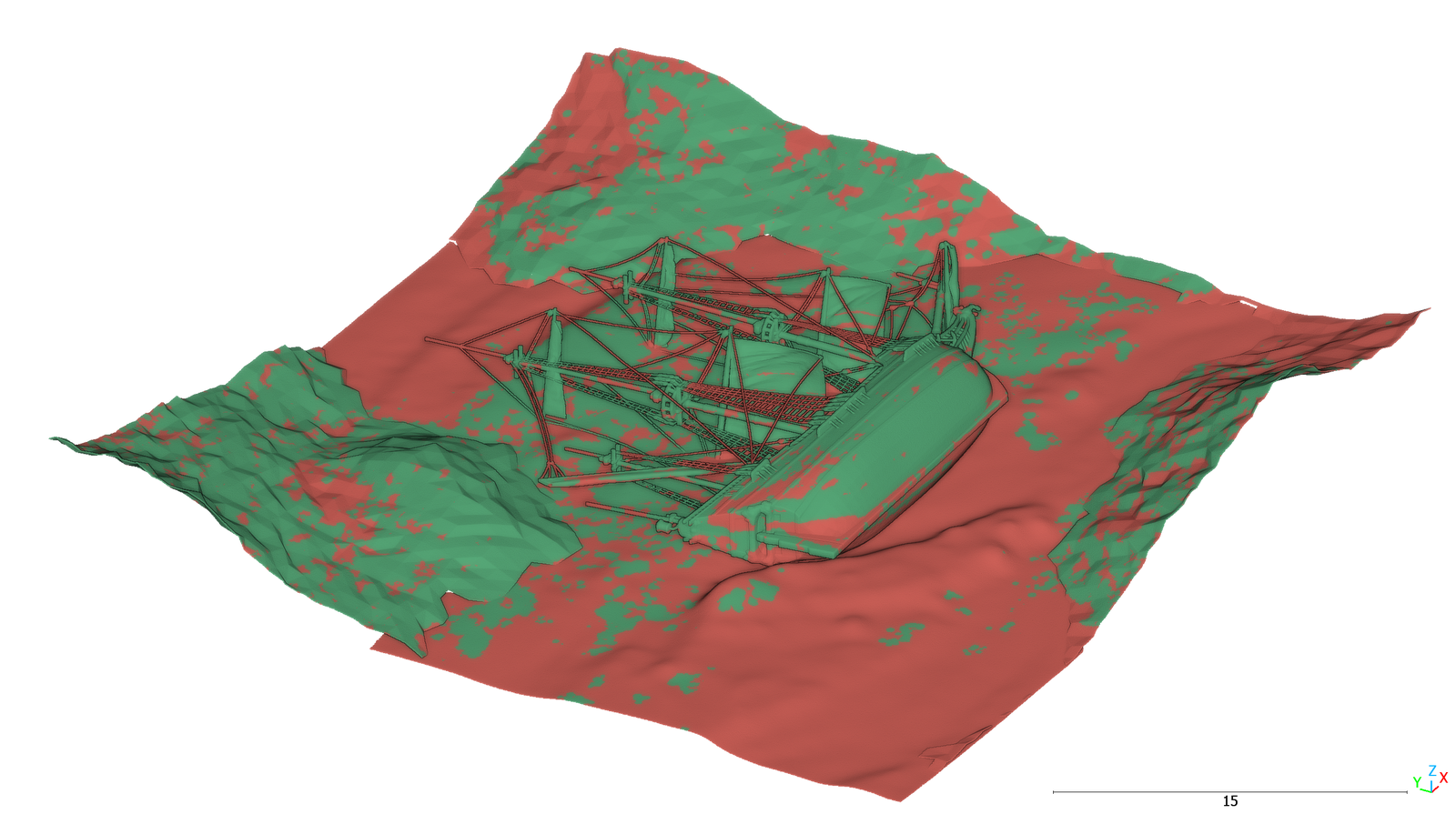} &
        \includegraphics[width=0.43\textwidth]{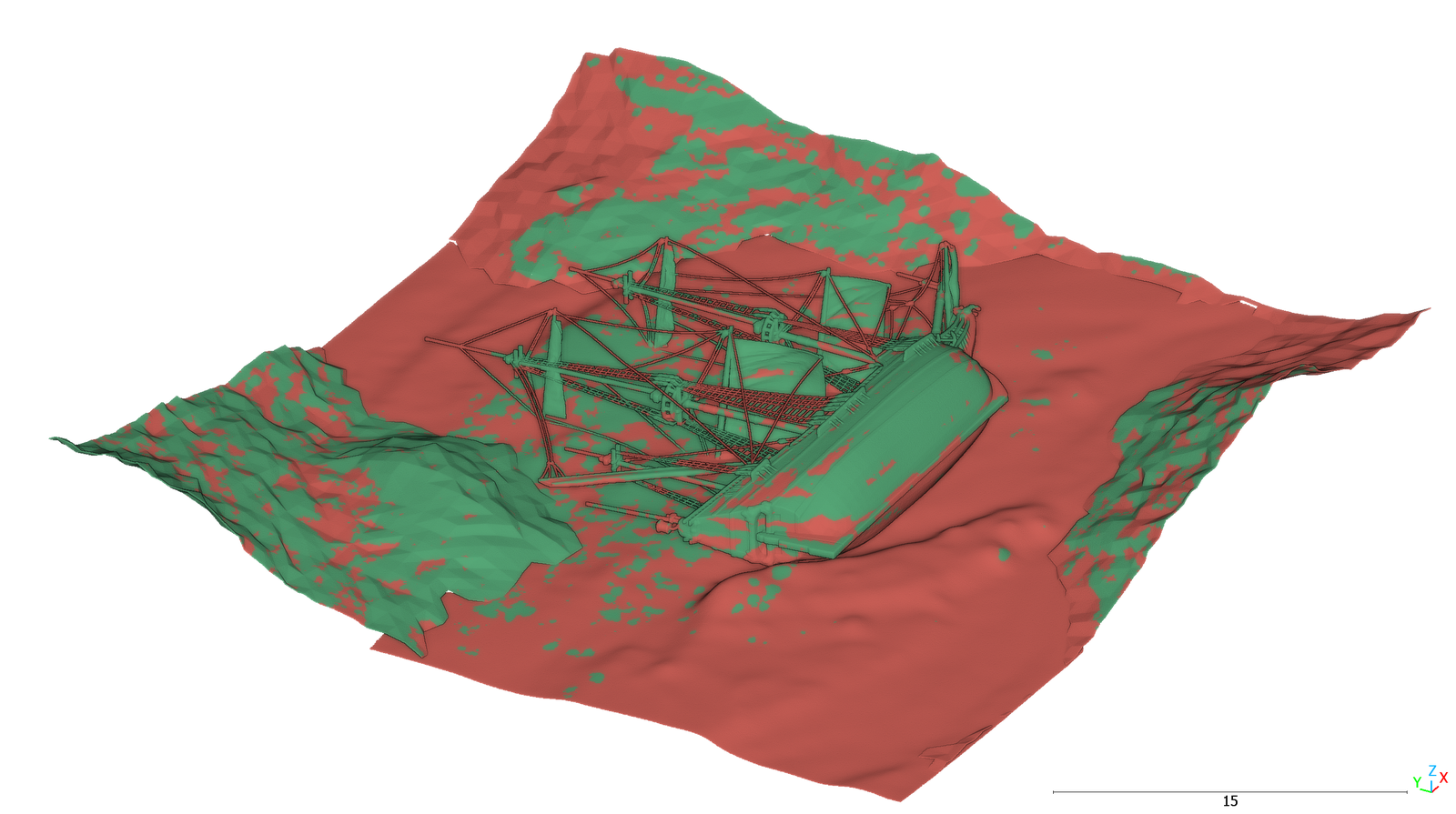} \\[6pt]
        \small\textbf{BathyFacto (Refr.\ ON)} & \small\textbf{MVS (refr.-corrected)} \\[3pt]
        \includegraphics[width=0.43\textwidth]{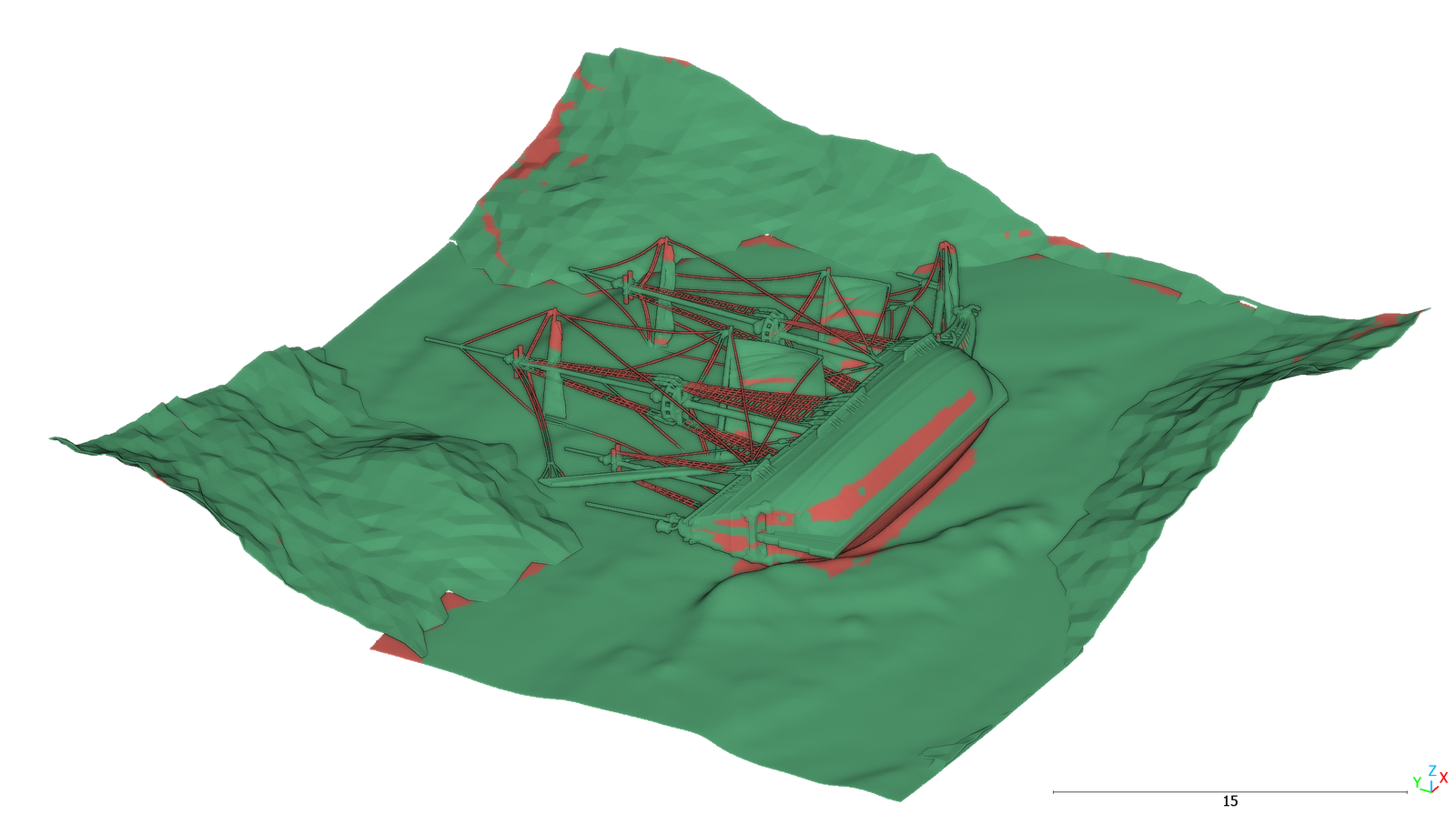} &
        \includegraphics[width=0.43\textwidth]{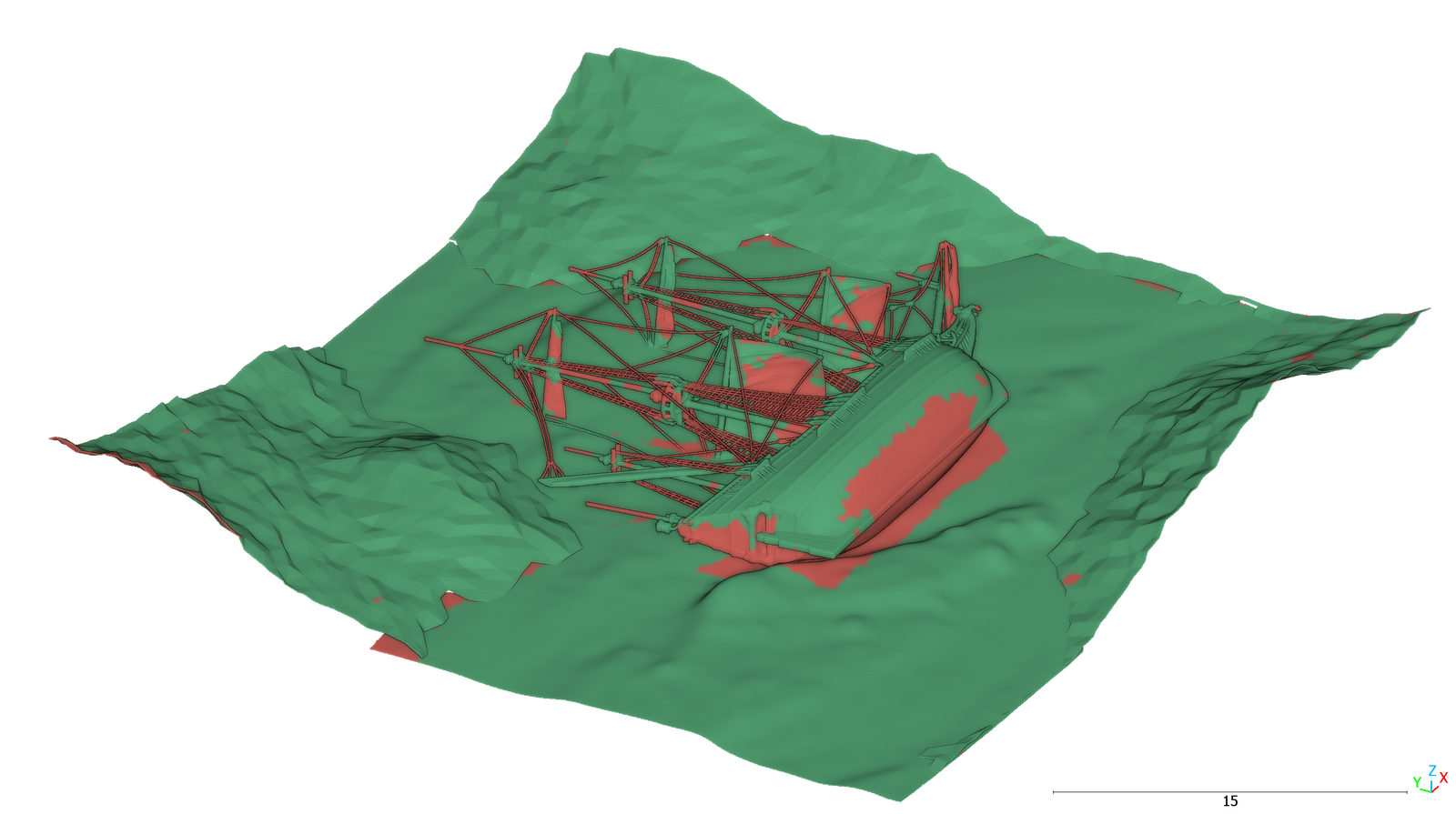} \\
    \end{tabular}

    \caption{Comparison of all four configurations: non-refracted NeRF baselines with approximate $+1/3$-depth correction, refraction-corrected MVS reconstruction, and BathyFacto (Refr.\ ON) as-is. \textbf{(a)}~C2M signed distance histograms; the histogram color fill follows the shared color scale of \autoref{fig:3d-evaluation-page}. Uncorrected distributions are identical in shape but shifted by the systematic refraction bias (Table~\ref{tab:quantitative-3d-results}). \textbf{(b)}~Completeness maps: green points lie within $0.2$\,m of the reference mesh, red points outside.}
    \label{fig:3d-fair-comparison}
\end{figure*}

\begin{figure*}[t]
    \centering
    \includegraphics[width=0.92\textwidth]{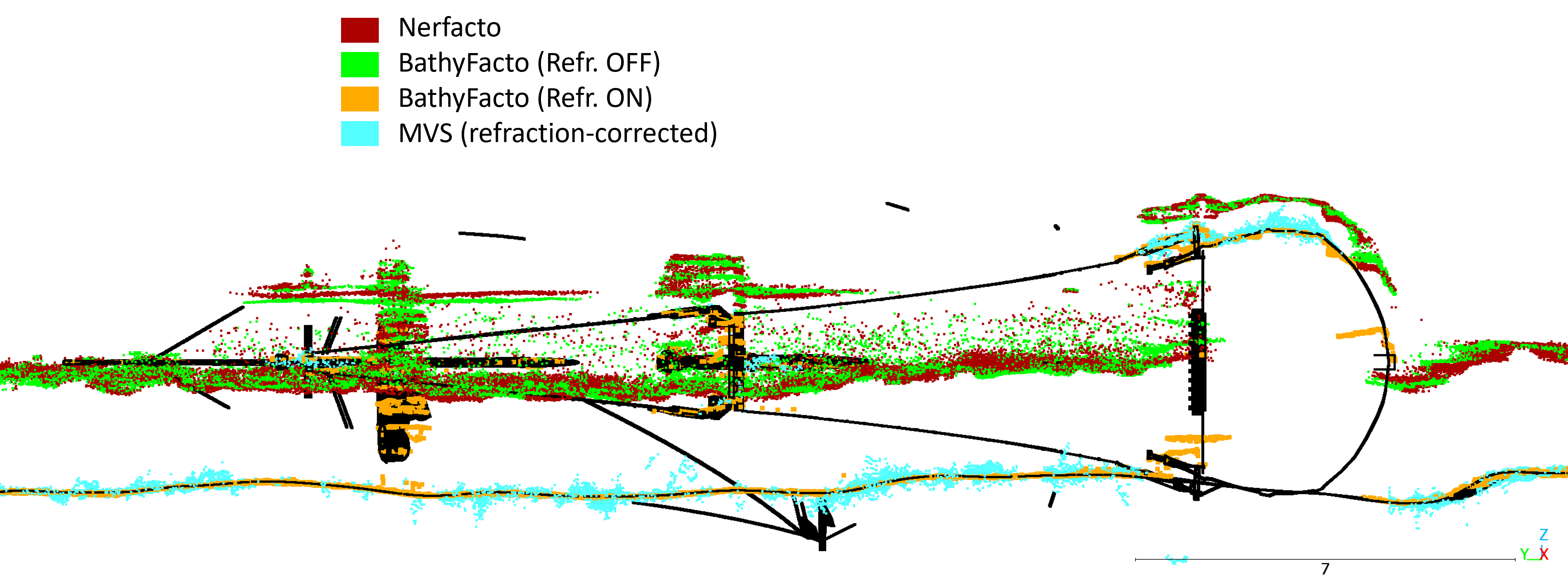}
    \caption{Cross-section through the ship hull. The refraction-enabled BathyFacto variant (orange) closely follows the reference geometry, whereas non-refracted variants exhibit a systematic depth error. The NeRF baselines are shown uncorrected here; their approximative $+1/3$-depth variants are reported in Table~\ref{tab:quantitative-3d-results} and \autoref{fig:3d-evaluation-page}. The MVS cloud is the refraction-corrected reconstruction, i.e.\ the complete refraction-aware photogrammetric workflow.}
    \label{fig:ship-cross-section}
\end{figure*}

To directly evaluate geometric reconstruction quality, we compare the exported point clouds against the reference mesh in the common global frame. Cloud-to-Mesh (C2M) signed distances, completeness, and a cross-section through the ship hull are reported in \autoref{fig:3d-evaluation-page}, \autoref{fig:3d-fair-comparison}, \autoref{fig:ship-cross-section}, and Table~\ref{tab:quantitative-3d-results}.

All variants are evaluated in the absolute global coordinate frame without any rigid-body (ICP)~\citep{besl_1992} alignment. For BathyFacto (Refr.\ ON), the reported C2M of $-0.001$\,m confirms that the refraction-aware model precisely recovers the 3D geometry without post-hoc registration. The non-refracted NeRF baselines (Nerfacto, BathyFacto Refr.\ OFF) exhibit a systematic depth offset from the missing refraction correction. For the Nerfacto and BathyFacto (Refr.\ OFF) variants, we therefore multiply the water depths by the refractive index of water ($n=1.333$). This simple physical refraction correction compensates the systematic depth underestimation and enables a fair comparison.

\begin{table*}[t]
\centering
\caption{Quantitative 3D point-cloud evaluation in the common reference frame, on the central
$44\times42$\,m$^2$ crop. C2M signed median, mean, and standard deviation are reported in the
absolute global coordinate frame without any rigid-body (ICP) alignment. BathyFacto (Refr.\ ON)
is reported as-is; for the non-refracted baselines (Nerfacto, BathyFacto Refr.\ OFF) and the MVS
reference, both the uncorrected result and a refraction-corrected variant are reported\textsuperscript{b}.
Completeness: percentage of reference-mesh points within $0.2$\,m of the nearest reconstructed
point \citep{hermann2024usegeo}.}
\label{tab:quantitative-3d-results}
\begin{threeparttable}
\small
\setlength{\tabcolsep}{4.8pt}
\renewcommand{\arraystretch}{1.1}
\begin{tabular}{@{}
    >{\raggedright\arraybackslash}p{0.29\textwidth}
    >{\raggedright\arraybackslash}p{0.19\textwidth}
    >{\centering\arraybackslash}p{0.12\textwidth}
    >{\centering\arraybackslash}p{0.11\textwidth}
    >{\centering\arraybackslash}p{0.09\textwidth}
    >{\centering\arraybackslash}p{0.10\textwidth}@{}}
\toprule
\textbf{Method} & \textbf{Correction} & \textbf{C2M Med. [m]} & \textbf{Mean [m]} & \textbf{Std [m]} & \textbf{Compl. [\%]} \\
\midrule
BathyFacto (Refr.\ ON)\tnote{a}       & ---              & $\mathbf{-0.001}$ & $\mathbf{-0.001}$ & $\mathbf{0.053}$ & \textbf{85.7} \\
\midrule
BathyFacto (Refr.\ OFF)               & none             & $+1.409$ & $+1.256$ & $0.880$ & 9.9 \\
BathyFacto (Refr.\ OFF)\tnote{b}      & $+1/3$ depth     & $+0.414$ & $+0.454$ & $0.330$ & 35.5 \\
Nerfacto                              & none             & $+1.370$ & $+1.219$ & $0.914$ & 11.6 \\
Nerfacto\tnote{b}                     & $+1/3$ depth     & $+0.406$ & $+0.437$ & $0.345$ & 44.4 \\
MVS (pairs, $\leq 15^\circ$)\tnote{c} & none (apparent)  & $+1.313$ & $+1.144$ & $0.694$ & 2.1 \\
MVS (pairs, $\leq 15^\circ$)\tnote{c} & refraction corr.\ & $+0.006$ & $+0.008$ & $0.064$ & 80.3 \\
\bottomrule
\end{tabular}
\begin{tablenotes}
\item[a] Evaluated in the absolute global coordinate frame without any rigid-body alignment of any kind.
\item[b] $+1/3$-depth correction: underwater points are deepened by the refractive-index factor
$n=1.333$ (nadir approximation), a simple physical refraction correction used in place of ICP.
\item[c] Refraction-corrected MVS (workflow in Sect.~\ref{sec:experimental-setup:configs}), restricted to
near-nadir camera groups ($\leq 15^\circ$); see Table~\ref{tab:mvs-sweep} for the coverage--accuracy trade-off.
\end{tablenotes}
\end{threeparttable}
\end{table*}

\begin{table}[t]
\centering
\begin{threeparttable}
\caption{Refraction-corrected MVS accuracy as a function of the maximum camera incidence angle of
the contributing image groups. Near-nadir groups are most accurate; including more oblique groups
raises coverage but degrades accuracy (increasing standard deviation), because the refraction
correction assumes near-nadir viewing. Completeness at $0.2$\,m on the central crop.}
\label{tab:mvs-sweep}
\small
\setlength{\tabcolsep}{4pt}
\renewcommand{\arraystretch}{1.1}
\begin{tabular}{@{}llccc@{}}
\toprule
\textbf{Groups} & \textbf{Max.\ tilt} & \textbf{C2M Med. [m]} & \textbf{Std [m]} & \textbf{Compl. [\%]} \\
\midrule
Pairs    & $\leq 15^\circ$          & $+0.006$ & $0.064$ & 80.3 \\
Pairs    & $\leq 30^\circ$          & $+0.007$ & $0.073$ & 81.2 \\
Pairs    & $\leq 63^\circ$          & $+0.011$ & $0.244$ & 89.8 \\
Triplets & $\leq 15^\circ$\tnote{a} & $-0.011$ & $0.188$ & 84.4 \\
Triplets & $\leq 63^\circ$          & $-0.014$ & $0.703$ & 93.2 \\
\bottomrule
\end{tabular}
\begin{tablenotes}
\item[a] The $\leq 30^\circ$ triplet set is identical to the $\leq 15^\circ$ set (no additional
groups satisfy the wider threshold).
\end{tablenotes}
\end{threeparttable}
\end{table}

\subsection{Ablation Analysis}
The 3D evaluation (Table~\ref{tab:quantitative-3d-results}) reveals distinct effects of refraction modeling:

\textbf{Refraction modeling.} BathyFacto with refraction enabled achieves a C2M signed median of $-0.001$\,m and the highest completeness (85.7\,\% at 0.2\,m tolerance), measured in the absolute global coordinate frame without any further alignment. This indicates that explicit Snell's-law refraction at the interface produces geometrically consistent and complete underwater point clouds. The BathyFacto no-refraction ablation shows a substantially larger C2M signed median ($+1.409$\,m) and lower completeness (9.9\,\%), confirming that the straight-ray model introduces a systematic depth offset and fails to reconstruct large portions of the underwater scene.

\textbf{Single-medium baseline.} The Nerfacto baseline (C2M signed median $+1.370$\,m, completeness 11.6\,\%) demonstrates that the single-medium model suffers from systematic depth bias without explicit refraction modeling.

\textbf{Comparison to refraction-corrected MVS.} The refraction-corrected MVS reference reaches a C2M signed median of $+0.006$\,m and 80.3\,\% completeness for the near-nadir image groups (Table~\ref{tab:quantitative-3d-results}), comparable to BathyFacto on the central crop. However, the per-point refraction correction is reliable only for near-nadir viewing. Table~\ref{tab:mvs-sweep} shows that including more oblique image groups raises coverage but degrades accuracy, with the standard deviation of the image pairs increasing from $0.06$\,m at $\leq 15^\circ$ to $0.24$\,m at $\leq 63^\circ$. With the present acquisition geometry, the near-nadir groups cover only the central scene region, so the corrected-MVS reference can be evaluated only there. BathyFacto, by contrast, models refraction per ray and reconstructs the full scene independently of the camera incidence angle, providing complete coverage where the corrected-MVS workflow cannot.

The cross-section through the ship hull (\autoref{fig:ship-cross-section}) visually confirms that refraction-enabled BathyFacto traces the reference geometry more faithfully below the water surface.

\subsection{Sensitivity to Interface Accuracy}
\label{sec:plane-sensitivity}
Because the water surface is modeled as a single tilted plane, the reconstruction accuracy depends on the accuracy of the estimated interface. To quantify this dependence, we systematically perturbed the fitted water plane and re-evaluated the BathyFacto (Refr.\ ON) reconstruction against the reference mesh. Three tilt perturbations rotate the plane normal by $1^\circ$, $2^\circ$, and $5^\circ$; three vertical offsets shift the plane by $+5$, $+15$, and $+30$\,cm. Table~\ref{tab:plane-sensitivity} reports the resulting C2M signed median and completeness; the unperturbed reference is $-0.001$\,m / 85.7\,\%.

\begin{table}[t]
\centering
\caption{Sensitivity of the BathyFacto (Refr.\ ON) reconstruction to water-plane perturbations, evaluated against the reference mesh on the central crop (completeness at $0.2$\,m). The unperturbed reference is $-0.001$\,m / 85.7\,\%.}
\label{tab:plane-sensitivity}
\small
\setlength{\tabcolsep}{6pt}
\renewcommand{\arraystretch}{1.1}
\begin{tabular}{@{}lcc@{}}
\toprule
\textbf{Perturbation} & \textbf{C2M Med. [m]} & \textbf{Compl. [\%]} \\
\midrule
Tilt $1^\circ$   & $-0.003$ & 81.8 \\
Tilt $2^\circ$   & $+0.008$ & 60.7 \\
Tilt $5^\circ$   & $+0.053$ & 33.4 \\
Offset $+5$\,cm  & $+0.027$ & 85.3 \\
Offset $+15$\,cm & $+0.083$ & 82.8 \\
Offset $+30$\,cm & $+0.152$ & 75.4 \\
\bottomrule
\end{tabular}
\end{table}

The reconstruction is robust to small tilt errors ($\leq 2^\circ$ keeps the C2M signed median below $1$\,cm) and to vertical offsets up to $15$\,cm (median below $9$\,cm, completeness above $82\,\%$). Tilt errors dominate over height errors: a $5^\circ$ tilt already lowers completeness to $33\,\%$, whereas a $30$\,cm vertical offset still retains $75\,\%$. The required accuracy --- a plane normal within a few degrees and a level within about $10$\,cm --- is attainable from standard UAV survey control points, indicating that the planar-interface assumption is practical whenever the surface is approximately flat.

\section{Discussion}
\label{sec:discussion}

This section contextualizes BathyFacto in prior work and practical UAV bathymetry requirements. We summarize where explicit two-media modeling helps, discuss remaining limitations, and highlight implications for future datasets and modeling choices.

\subsection{Contextualization and Comparison to Prior Work}
BathyFacto builds on NeRFrac's foundational idea of embedding Snell's law into a NeRF pipeline \citep{zhan2023nerfrac} but differs in several key aspects. First, BathyFacto uses a shared hash-grid-based density field with a medium-conditioned color head rather than fully separate air and water fields or a jointly optimized interface shape. This shared-geometry design enforces geometric consistency across the air-water boundary and converges substantially faster than sinusoidal-encoding approaches (100K vs.\ 200K+ iterations). Second, BathyFacto employs a single proposal-network sampler with a kinked density wrapper, which enables adaptive sample allocation across the air--water boundary without maintaining separate sampling hierarchies. Third, BathyFacto integrates directly into the Nerfstudio framework, inheriting its proposal-network sampling, appearance embeddings, and camera pose optimization---features not available in NeRFrac's vanilla-NeRF backbone. Fourth, BathyFacto provides a complete pipeline from photogrammetric input to metrically consistent 3D point clouds on simulated data in the original coordinate frame, which prior refractive NeRF research has not demonstrated.

Compared to classical through-water photogrammetry corrections \citep{agrafiotis2019correcting, kastner2023iterative}, BathyFacto operates on the volumetric representation level rather than correcting individual rays post-hoc, enabling joint optimization of geometry, appearance, and camera poses within a single framework.

\subsection{Completeness evaluation}
The completeness maps (\autoref{fig:3d-fair-comparison}) visualize the spatial distribution of reconstructed points relative to the reference mesh. Green points lie within $0.2$\,m of the reference mesh; red points deviate by more than $0.2$\,m and are therefore counted as not reconstructed in the completeness metric. The non-refracted NeRF baselines use an approximate depth correction factor ($n=1.333$) of their underwater points. BathyFacto with refraction enabled achieves 85.7\,\% completeness in the absolute global frame without any alignment, substantially outperforming both non-refracted baselines even after their $1/3$-depth correction (35.5\,\% for the no-refraction ablation and 44.4\,\% for Nerfacto).

\subsection{Strengths and Limitations}

\textbf{Strengths.} (i) The shared-geometry architecture with hash-grid encoding provides fast convergence and geometrically consistent two-media reconstruction. (ii) Nerfstudio integration ensures reproducibility and access to ongoing framework improvements. (iii) The refraction-corrected point cloud export uses the original photogrammetric camera poses and reversible coordinate transforms, enabling direct metric comparison with reference geometry in the original coordinate frame.

\textbf{Limitations.} (i) The water surface must be known a priori and is currently modeled as a plane; wavy or time-varying surfaces are not yet supported. (ii) Medium masks must be provided or generated in a preprocessing step; mask errors at shorelines can cause artifacts. (iii) The model has been validated only on synthetic data; real-world UAV imagery introduces additional challenges such as turbidity, caustics, and specular reflections. Additionally, the model uses a fixed index of refraction $n = 1.333$, which is appropriate for the synthetic dataset but ignores the variation of $n$ with water temperature, salinity, and wavelength in real deployments --- a limitation that should be addressed when applying the method to field data. (iv) The evaluation uses a single synthetic scene with one seabed texture; systematic validation across several scenes spanning different depths, bottom textures, and bed slopes --- including acquisition of a real-world dataset --- remains future work.

The method relies on manually defined binary water masks to separate the two-media segments. In practical UAV applications, shoreline boundaries can be ambiguous due to shallow submerged structures, riparian vegetation, turbidity gradients, and shadows. The current implementation treats mask errors as hard failures, but the architecture is compatible with soft, probabilistic masks --- an extension we identify as future work. Jointly learned masks, such as those produced by semantic segmentation networks, pre-trained on multispectral imagery, represent a promising direction for automating the segmentation step. In practice, the water area is not segmented individually in each image but rather, it is defined once in object space based on the photogrammetric surface model and the adjusted water surface, and then back-projected onto each image. Overhanging riparian vegetation is then resolved by occlusion testing against the photogrammetric surface model (DSM/DOM), while the shoreline results from the intersection of the water plane with the terrain model. This object-space approach is expected to yield geometrically consistent masks across all views and should be more robust than per-image 2D segmentation at ambiguous shorelines.

\textbf{Anticipated real-data failure modes.} Beyond the limitations above, several effects are expected to challenge real UAV deployments. \emph{Turbidity} attenuates radiance non-uniformly with depth and reduces the effective depth limit; the optional Beer--Lambert absorption term can partly account for it but cannot recover signal lost to strong scattering. \emph{Sun glint} produces high-radiance, view-dependent specular highlights at the air--water interface that a Lambertian two-media appearance model cannot explain as seafloor colour, which can bias the reconstructed surface density. In this case, an explicit reflection channel is a promising mitigation and a direction for future work. \emph{Motion blur} and \emph{imperfect calibration} degrade NeRF reconstruction in general and call for careful image selection and calibration validation. Real-data experiments at the Almsee lake and the Pielach river, evaluated against airborne laser bathymetry, are planned as future work.

\section{Conclusion and Outlook}
\label{sec:conclusion}

We have presented BathyFacto, a refraction-aware two-media extension of Nerfacto for photogrammetric bathymetry. BathyFacto uses a shared hash-grid-based density field with a medium-conditioned color head to represent air and water regions, traces camera rays as two piecewise linear segments connected by Snell's-law refraction at a planar water surface, and employs a single proposal-network sampler with a kinked density wrapper to allocate samples adaptively across both media. Integrated into Nerfstudio, the method provides a complete pipeline from photogrammetric SfM output to metrically consistent 3D point clouds on simulated data in the original coordinate frame.

On a synthetic two-media scene with known ground truth, BathyFacto with refraction enabled achieves 85.7\,\% completeness at 0.2\,m tolerance and a C2M signed median of $-0.001$\,m, measured in the absolute global coordinate frame without any alignment. This substantially outperforms both the single-medium Nerfacto baseline ($+1.370$\,m, 11.6\,\%) and the no-refraction ablation ($+1.409$\,m, 9.9\,\%). The ablation with refraction disabled confirms that explicit two-media ray modeling is the primary driver of improved geometric completeness.

Future work will address the following directions: (i) extending the planar interface model to wavy or time-varying water surfaces, which is critical for real-world river and coastal applications; (ii) validating the method on real UAV imagery with ground-truth bathymetric references from airborne laser bathymetry; (iii) optimizing the loss function specifically for bathymetric accuracy rather than image-space fidelity; and (iv) exploring the reconstruction of submerged vegetation (macrophytes) as a challenging test case for volumetric underwater reconstruction.

\section*{Acknowledgements}
The authors thank the Nerfstudio team for their open-source framework. Computations were performed on the GPU infrastructure of TU Wien. The BathyFacto source code and training configurations will be released as open-source software (Apache-2.0 License), based on the Nerfstudio framework, in a dedicated public repository upon acceptance, with a persistent DOI assigned via Zenodo upon release. The authors acknowledge TU Wien Bibliothek for financial support through its Open Access Funding Programme.

\section*{Funding}
The presented research was carried out within the BathyNeRF project. This research was funded in whole or in part by the German Research Foundation (DFG, 538522540) and the Austrian Science Fund (FWF) [10.55776/PIN1353223]. For open access purposes, the author has applied a CC BY public copyright license to any author accepted manuscript version arising from this submission.

\section*{Declaration of Competing Interest}
The authors declare that they have no known competing financial interests or personal relationships that could have appeared to influence the research reported in this contribution.

\printcredits
\bibliographystyle{cas-model2-names}
\bibliography{references}

\end{document}